\documentclass[lettersize,journal]{IEEEtran}
\usepackage{amsmath,amsfonts}
\usepackage{algorithmic}
\usepackage{algorithm}
\usepackage{array}
\usepackage{textcomp}
\usepackage{stfloats}
\usepackage{url}
\usepackage{verbatim}
\usepackage{graphicx}
\usepackage{cite}
\usepackage{booktabs}
\usepackage{arydshln}
\usepackage{xcolor}
\usepackage{subcaption}
\usepackage{amsmath}
\usepackage[colorlinks=true]{hyperref}
\usepackage{cleveref}
\usepackage{amssymb}
\usepackage{multirow}
\begin{document}

\title{Multi-label Guided Soft Contrastive Learning for Efficient Earth Observation Pretraining}

\author{Yi Wang,~\IEEEmembership{Student Member,~IEEE}, Conrad M Albrecht,~\IEEEmembership{Member,~IEEE}, Xiao Xiang Zhu,~\IEEEmembership{Fellow,~IEEE}
\thanks{Y. Wang (yi4.wang@tum.de) and X. X. Zhu (xiaoxiang.zhu@tum.de) are with the Chair of Data Science in Earth Observation, Technical University of
Munich (TUM). X. X. Zhu is also with the Munich Center for Machine Learning.  C. M. Albrecht (conrad.albrecht@dlr.de) is with the Remote Sensing Technology Institute, German Aerospace Center (DLR). }
}



\maketitle

\begin{abstract}
Self-supervised pretraining on large-scale satellite data has raised great interest in building Earth observation (EO) foundation models. However, many important resources beyond pure satellite imagery, such as land-cover-land-use products that provide free global semantic information, as well as vision foundation models that hold strong knowledge of the natural world, are not widely studied. In this work, we show these free additional resources not only help resolve common contrastive learning bottlenecks, but also significantly boost the efficiency and effectiveness of EO pretraining.

Specifically, we first propose soft contrastive learning that optimizes cross-scene soft similarity based on land-cover-generated multi-label supervision, naturally solving the issue of multiple positive samples and too strict positive matching in complex scenes. Second, we revisit and explore cross-domain continual pretraining for both multispectral and SAR imagery, building efficient EO foundation models from strongest vision models such as DINOv2. Adapting simple weight-initialization and Siamese masking strategies into our soft contrastive learning framework, we demonstrate impressive continual pretraining performance even when the input modalities are not aligned. 

Without prohibitive training, we produce multispectral and SAR foundation models that achieve significantly better results in 10 out of 11 downstream tasks than most existing SOTA models. For example, our ResNet50/ViT-S achieve 84.8/85.0 linear probing mAP scores on BigEarthNet-10\% which are better than most existing ViT-L models; under the same setting, our ViT-B sets a new record of 86.8 in multispectral, and 82.5 in SAR, the latter even better than many multispectral models. Dataset and models are available at \url{https://github.com/zhu-xlab/softcon}. 


\end{abstract}


\section{Introduction}
\IEEEPARstart{S}{elf-supervised} learning has driven wide attention in pretraining Earth observation (EO) foundation models on large-scale satellite data~\cite{wang2022self,tao2023self,zhu2024foundations}. While more and more efforts are spent on scaling up the data and model size with purely unsupervised pretraining, many other resources such as various land cover land use products tend to be overlooked. For example, ESA WorldCover~\cite{zanaga_2021_5571936,zanaga_2022_7254221} provides the first global land cover maps for 2020 and 2021 at 10 m resolution, and Google Dynamic World~\cite{brown2022dynamic} provides a continuous dataset of near-real-time land use land cover mapping. These dense products are highly correlated with commonly studied medium-resolution satellite imagery, and offer free semantic annotations with real-global coverage. Even though they are noisy at pixel-level due to semi-automatic creation process, they can be easily integrated into scene-level annotations with rather good quality. A similar concept has been demonstrated valid in supervised pretraining in GeoKR \cite{li2021geographical}, where land cover products and geographical location are regarded as geographical knowledge to provide supervision. In this work, we will show the benefits of such auxiliary information in extending the popular contrastive self-supervised learning framework to build EO foundation models.

\begin{figure}[t]
    \centering
    \includegraphics[width=\columnwidth]{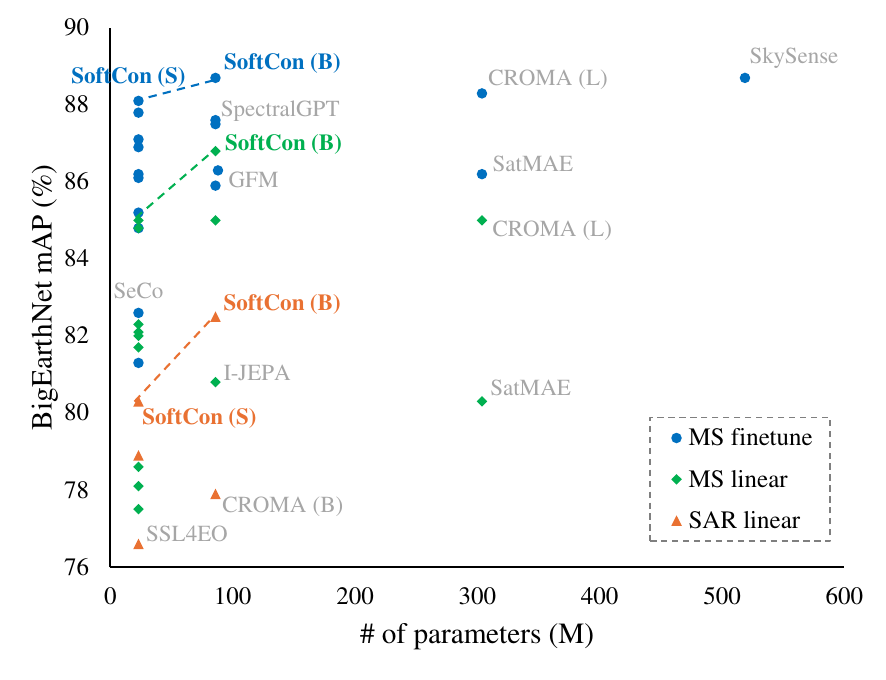}
    \caption{A visual comparison of transfer learning performances on BigEarthNet-10\%. SoftCon (ours) achieves SOTA results with lighter backbones on both linear probing and fine-tuning, and both multispectral and SAR. Our best multispectral linear result is comparable to best models' fine-tuning; our best SAR result outperforms many multispectral models. See \Cref{fig:scatter-split} for separated views of each setting.}
    \label{fig:be-efficiency}
\end{figure}

Contrastive learning with negative sampling such as SimCLR~\cite{chen2020simple} and MoCo~\cite{he2020momentum} have been shown robust and effective in EO pretraining \cite{wang2022self,wang2023ssl4eo}. These methods pull together features of augmented views from the same anchor image (as positive sample), and push away features of other images (as negative samples). The negative samples are usually from a batch or a memory bank, and the model is trained to identify the positive sample from the negative ones. While such instance-discrimination-based methods learn good representations and have been widely used, they inevitably bear the risk of discriminating false negative samples. Specifically, in a large negative pool, there are likely samples that are not the same as the anchor, but have very similar semantic information (e.g., belong to the same class). This issue is more significant in EO compared to natural images, as the Earth has a limited surface and the landscapes are usually very redundant.

Several methods have been proposed in computer vision to solve the false negative conflicts, of which one representative is supervised contrastive learning (SupCon)~\cite{khosla2020supervised}. Leveraging image labels, SupCon defines positive samples as images belonging to the same semantic class. During training, samples from the same class are pulled together in the embedding space, while those from different classes are pushed away. Such a simple multi-positive design has been proven beneficial in natural images, and as we mentioned earlier, bears great potential to leverage the free land-use-land-cover annotations in EO. However, a satellite image usually contains more complex semantics than a single class, leading to at least multi-label annotations. Although one can define positive samples as perfect matching of each label component, this would result in two similar images with slightly different label distributions being forced apart (e.g. two neighboring urban scenes one with a small part of river and the other not). 

To solve this problem, we extend SupCon to the multi-label scenario, proposing a novel soft contrastive learning method (we term as SoftCon) that takes into account the similarity of complex scenes. Specifically, we calculate cosine similarities of the normalized multi-label one-hot vectors across different scenes. Images with identical labels thus have the highest similarity scores, and images with similar labels have higher scores than images with very different labels. Then, we train the model to directly learn such cross-scene similarities by optimizing a soft contrastive loss on the cosine similarities of corresponding feature projections. In this way, semantically more similar images are pulled closer than semantically more dissimilar images, ending up with a smoothly distributed latent space. To prepare the training data, we match SSL4EO-S12 \cite{wang2023ssl4eo} images with Dynamic World \cite{brown2022dynamic} segmentation maps and integrate scene-level multi-label annotations, building a large-scale global multi-label classification dataset.

Meanwhile, another important resource that has huge potential in helping build EO foundation models lies in the general computer vision community: the vision foundation models. These models are exhaustively trained on huge amount of natural images, and have already gained strong knowledge of the visual world. Dating back to before the era of foundation models, ImageNet pretrained weights had been widely used and proved beneficial in many supervised Earth observation tasks. Similarly, they can also be used in unsupervised learning, leading to cross-domain continual pretraining. In this regard, recent works such as GFM \cite{mendieta2023towards} propose to build EO foundation models by distilling frozen ImageNet models. However, GFM-style training is limited to RGB images, restricting the flexibility to adapt to various EO sensors. Furthermore, the fast advances in computer vision have made available much stronger vision models than ImageNet supervised weights. To bridge this gap, we revisit the natural idea of simple weight initialization which has been preliminarily explored and verified effective in recent works \cite{wang2024samrs,wang2024mtp}. For unaligned input modalities, we simply leave the first layer's weights randomly initialized. In addition, to save hardware memory when continually training large Vision Transformers, we adopt Siamese masking inspired by masked Autoencoder \cite{he2022masked}. Specifically, we randomly mask out a certain percentage of input patches on the trainable branch of the Siamese contrastive learning framework, and only send the remaining visible patches to the encoder. We show such simple but flexible continual pretraining strategies, when applied with strong vision foundation models such as DINOv2 \cite{oquab2023dinov2}, exhibit impressive effectiveness and efficiency even when the imaging sensors are completely different from the source domain. 

Integrating the continual pretraining strategies into the soft contrastive learning framework, we efficiently train CNN and ViT foundation models that reach SOTA performances in 10 out of 11 downstream tasks. For example, our ResNet50/ViT-S (23M parameters, 100 epochs) achieve 84.8/85.0 linear probing mAP scores on BigEarthNet-10\% which are better than most existing ViT-L models (300M parameters, 100-300 epochs) and comparable to CROMA and SkySense ($\ge$600 epochs); under the same setting, our ViT-B (86M parameters) sets a new SOTA of 86.8 in multispectral, and 82.5 in SAR, the latter even better than most existing multispectral models.

In summary, our contributions are as follows:
\begin{itemize}
    \item We explore the benefits of open resources beyond pure satellite imagery for efficient EO pretraining, producing multispectral and SAR foundation models that reach SOTA performances in 10 out of 11 downstream tasks.
    \item We propose soft contrastive learning that guides contrastive pretraining with land-cover-generated multi-label supervision. 
    \item As a side product, we release a global multi-label scene classification dataset by matching noisy Dynamic World land cover maps with SSL4EO-S12 images and integrating good-quality multi-label annotations.
    \item We revisit cross-domain continual pretraining with simple weight initialization from strong vision foundation models and Siamese masking. We verify that even when the input modalities are not aligned, the strong knowledge can still be efficiently transferred to the target EO domain.
\end{itemize}

\section{Related Work}

\subsection{Earth observation foundation models}
Massive research has been conducted on the development of Earth observation (EO) foundation models. While there is also a line of supervised pretraining \cite{wang2022empirical} on large-scale labeled datasets \cite{long2021creating,zhang2022artificial}, a majority of works tackle the technical adaptation of self-supervised pretraining techniques into EO domain. Early works focus on EO-specific data characteristics for contrastive view generation. For example, Tile2vec \cite{jean2019tile2vec} proposed to pull together geospatially close tiles while pushing away far tiles. SeCo \cite{manas2021seasonal} proposed to use different seasons as augmented views for contrastive learning. CACo \cite{mall2023change} proposed to perceive temporal changes with the spatiotemporal structure of remote sensing time series. Further, another group of works explores masked-image-modeling-based pretraining \cite{wang2022advancing}. SatMAE \cite{cong2022satmae} proposed temporal and spectral positional encoding for multispectral imagery and time series. RingMo \cite{sun2022ringmo} proposed less aggressive masking. Scale-MAE \cite{reed2023scale} proposed GSD-based positional encoding and multi-scale reconstruction. SpectralGPT \cite{hong2023spectralgpt} proposed 3D masking to encode and reconstruct spectral data. FG-MAE \cite{wang2023feature} proposed to reconstruct remote sensing image features such as normalized difference indices. Going beyond a single modality and time stamp, CROMA \cite{fuller2024croma}, DeCUR \cite{wang2023decur}, OFA-Net \cite{xiong2024one} and DOFA \cite{xiong2024neural} investigated multi-modal multi-sensor pre-training, RemoteCLIP \cite{liu2024remoteclip}, SkyScript \cite{wang2024skyscript} and BITA \cite{yang2024bootstrapping} explored EO vision language pretraining, RingMo-sense \cite{yao2023ringmo}, Presto \cite{tseng2023lightweight} and Prithvi \cite{jakubik2023foundation} studied EO time series, and SkySense \cite{guo2023skysense} combined both modality and time sequence in a unified architecture, reaching SOTA performance in many downstream tasks. Meanwhile, another line of research targets the curation of EO pretraining datasets, such as SSL4EO-S12 \cite{wang2023ssl4eo} for Sentinel-1 and 2, SSL4EO-L \cite{stewart2024ssl4eo} for Landsat series, and SatlasPretrain \cite{bastani2023satlaspretrain} for medium- and high-resolution satellite and aerial imagery with extensive annotations. While a general trend is to scale the data and model sizes with exhaustive training cost, important existing resources such as open annotations and vision foundation models tend to be overlooked. In this study, we share the conceptual insight with GeoKR \cite{li2021geographical} to use land-cover-generated multilabel annotations to guide EO pretraining, and with GFM \cite{mendieta2023towards} and DOFA to utilize vision models for continual pretraining.

\begin{figure*}[]
    \centering
    \includegraphics[width=0.9\textwidth]{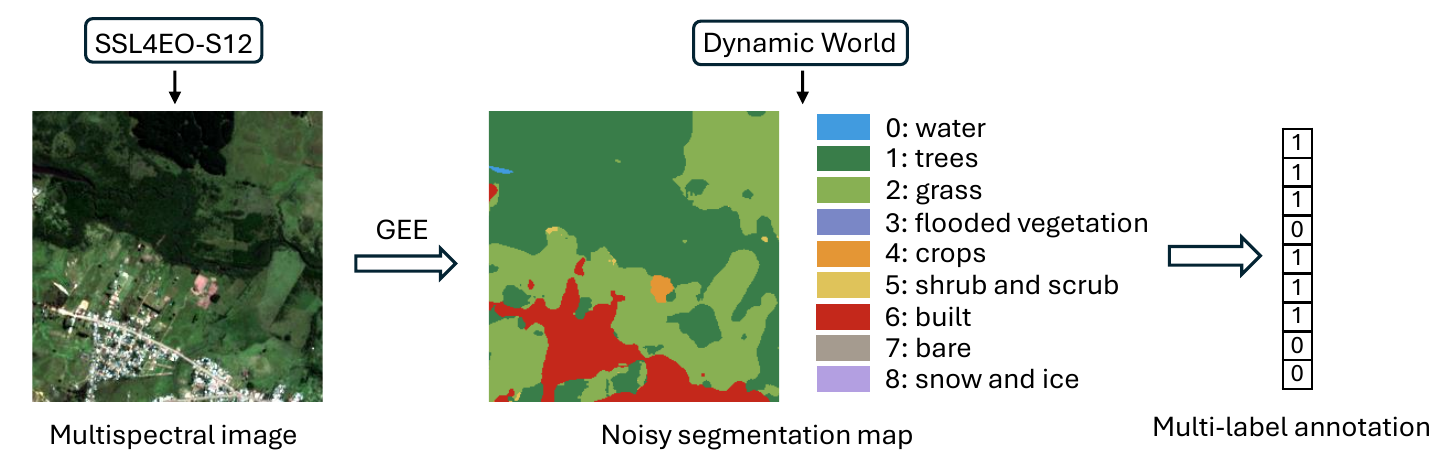}
    \caption{The workflow of the multi-label dataset curation.}
    \label{fig:dataset}
\end{figure*}

\subsection{Multi-positive contrastive learning}

Contrastive learning beyond simple instance discrimination has been widely explored in computer vision. SupCon \cite{khosla2020supervised} proposed to use image labels for positive matching, extending contrastive learning to the fully-supervised setting. Almost at the same time, ML-CPC \cite{song2020multi} proposed multi-label contrastive predictive coding, identifying multiple positive samples as a multi-label classification problem\footnote{We note that depending on the context, "multi-label" in this paper may refer to two things: 1) the contrastive learning objective with multi-positive matching; 2) the input images are multi-label annotated.}. Further on, WCL \cite{zheng2021weakly} combined contrastive instance discrimination with SupCon by predicting pseudo weak labels. Sel-CL \cite{li2022selective} extended SupCon to deal with noisy labels. All of these methods still deal with single-label datasets. To deal with more complex scenes, HiMulCon \cite{zhang2022use} presented a hierarchical representation learning framework that can leverage all available labels and preserve the hierarchical relationship
between classes. MLS \cite{zhu2023multi} proposed to assign multiple binary pseudo-labels for each input image by comparing its embeddings with those in two dictionaries, and training the model with binary cross-entropy loss. However, these two methods are still based on hard multi-positive matching, restricting the soft relationship between images with overlapped multi-label distributions. To tackle this issue, and to make the best use of real-world multi-label annotations, we propose soft contrastive learning that allows soft positive matching between different multi-label scenes.

\subsection{Continual pretraining}

Early works of continual pretraining were mainly developed in natural language processing to improve domain-specific large language models \cite{gururangan2020don,liu2021continual}. In vision, existing works like \cite{kalapos2022self} and \cite{reed2022self} proposed hierarchical pretraining approaches for task adaptation, while not targeting task-agnostic representations. In remote sensing, CSPT \cite{zhang2022consecutive} and TOV \cite{tao2023tov} proposed consecutive pretraining from natural images to remote sensing images, yet they are both limited to retraining on natural images for the source model. SpectralGPT \cite{hong2023spectralgpt} used a similar progressive pretraining pipeline between EO datasets to benefit from their unique advantages, yet it still requires first-stage pretraining from scratch. GFM \cite{mendieta2023towards} and DOFA \cite{xiong2024neural} explored continual pretraining for generic EO foundation models by distilling knowledge from frozen vision backbones. However, it heavily relies on ImageNet weights whose prior knowledge is restricted compared to stronger vision foundation models. Meanwhile, GFM-style continual pretraining can only process RGB data, limiting the flexibility to adapt to various EO modalities such as multispectral imagery. To bridge this gap, we revisit and explore the simple cross-domain continual pretraining with weight initialization \cite{wang2024samrs,wang2024mtp} from strong vision foundation models such as DINOv2 \cite{oquab2023dinov2}. For non-RGB imagery, we simply leave the input embedding layer randomly initialized. We show such a strategy, though naive, is both flexible and impressively effective in EO pretraining regardless of target sensors.

\section{Methodology}

\subsection{Building a large-scale multi-label dataset}

We first build a global multi-label land-cover-land-use classification dataset by automatically matching multispectral/SAR imagery from a large-scale satellite dataset with open land cover products and integrating pixel-level labels to scene-level multi-label annotations. 

We choose SSL4EO-S12 \cite{wang2023ssl4eo} as the source of satellite imagery, a multi-modal multi-temporal dataset specifically designed for large-scale self-supervised learning. The dataset consists of 4-seasonal Sentinel-1/2 SAR-optical images from 251,079 non-overlapping locations in the world, covering a wide range of geographical and temporal diversities. For accurate spatial and temporal matching of the scenes, we choose Dynamic World \cite{brown2022dynamic} to get the global land cover maps, which provide continuous land cover monitoring in 9 semantic classes. Both datasets are derived from Sentinel data stored in Google Earth Engine, and thus can be well aligned with the exact metainformation such as the acquisition time.

\Cref{fig:dataset} shows the general workflow of the curation of the multi-label dataset. Based on the geospatial coordinates and acquisition time, we match each SSL4EO-S12 L1C multispectral image with its corresponding Dynamic World land cover map in Google Earth Engine. Due to the effect of clouds, a few images do not have a corresponding segmentation map. This results in 247,377 locations on which there's a successful match for at least one season. Then, we gather the pixel labels into scene-level multi-labels for each image. As can be seen from the example scene in \Cref{fig:dataset}, though the segmentation map is noisy, the 
scene-level semantic classes are rather accurate.

\begin{figure*}[]
    \centering
    \includegraphics[width=\textwidth]{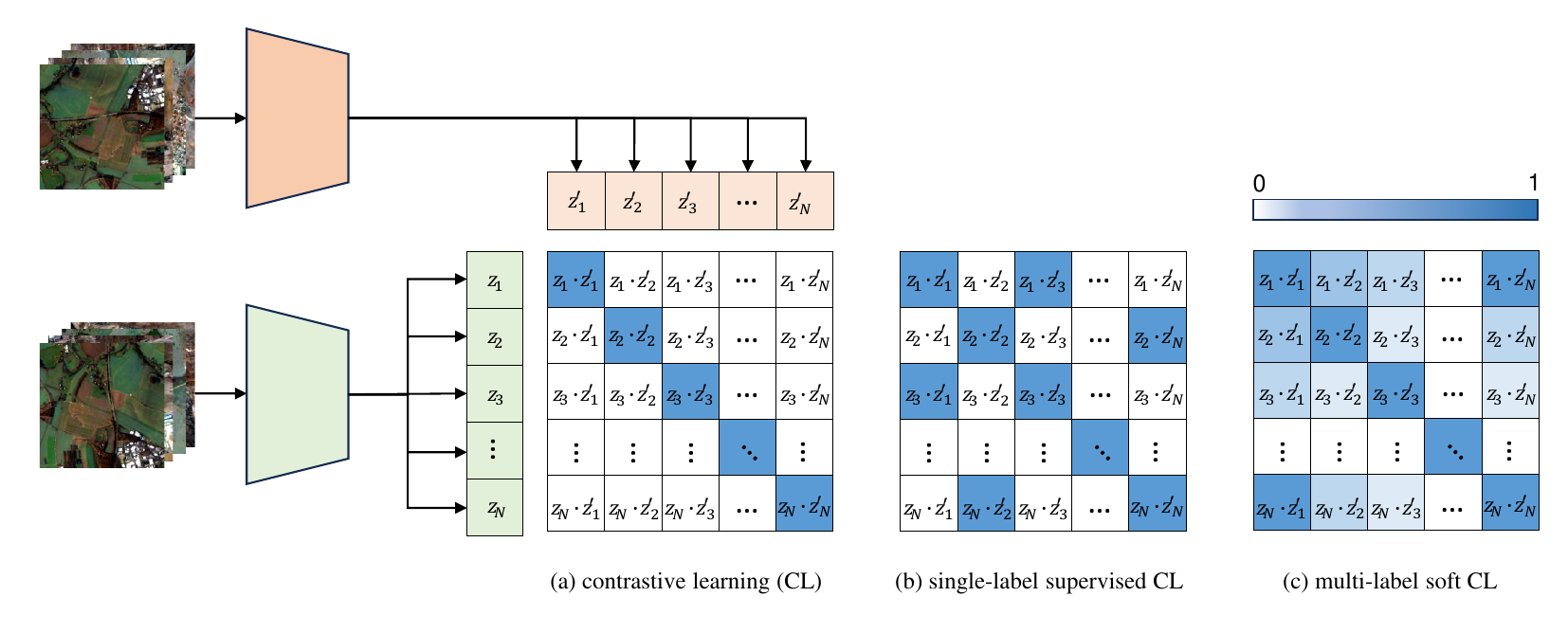}
    \caption{Different contrastive learning designs. (a) The original contrastive learning performs strict instance discrimination where one anchor image has only one positive pair; (b) supervised contrastive learning allows multiple positive responses when images belong to the same class; (c) our proposed soft contrastive learning can effectively exploit multi-label annotations by assigning soft similarity scores to different pairs. SofCon is a more generic design that covers SupCon: when multi-label degrades to single-label, the soft similarities turn into binary scores, thus becoming multi-positive supervised contrastive learning.}
    \label{fig:contrast}
\end{figure*}

In total, we get 780,371 annotated multi-label images, each with size 264$\times$264 in 10m resolution. We term this final dataset SSL4EO-S12-ML, and will release it for further research.
Notably, SSL4EO-S12-ML can also be used as a large-scale multi-label benchmark dataset that covers the whole globe, complementing existing datasets like BigEarthNet \cite{sumbul2019bigearthnet}. For better reference, we provide a comprehensive dataset sheet and supervised benchmarking results in the appendix. In addition, we note that SSL4EO-S12-ML labels are derived from Sentinel-2 multispectral images, which are spatially well aligned with the corresponding Sentinel-1 SAR images, but temporally there can be shifts. Therefore, the SAR version is generally less accurate than the multispectral version. Nevertheless, it is enough to to guide our pretraining.

\subsection{Multi-label guided soft contrastive learning}
\label{subsec:softcon}

In this section, we introduce soft contrastive learning (SoftCon), which improves upon the original contrastive learning and supervised contrastive learning (SupCon) \cite{khosla2020supervised} to effectively utilize multi-label annotations.

\begin{figure*}[]
    \centering
    \includegraphics[width=0.95\textwidth]{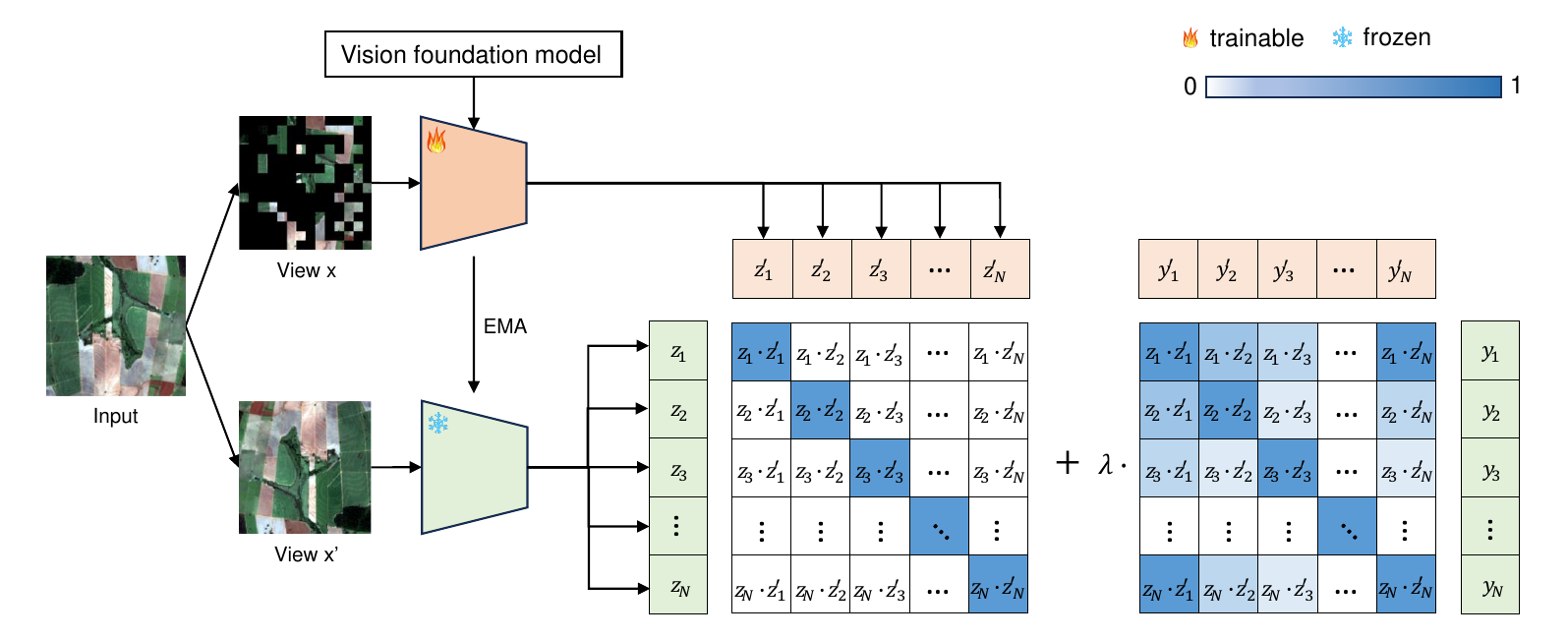}
    \caption{The general framework of SoftCon. Given a batch of input images, two batches of augmented views are parallelly sent through the two branches of a Siamese network. A similarity matrix is calculated based on the resulting two batches of feature vectors. A weighted sum of the contrastive and soft contrastive loss is optimized. For contrastive loss, this matrix should be close to Identity; for soft contrastive loss, this matrix should be close to the similarity matrix of the label vectors. We load vision foundation models for both the trainable (base encoder) and the frozen (momentum encoder) branches when initializing the model. During training, the weights of the momentum encoder are updated by exponential moving average (EMA) of the base encoder: $\theta_{\mathrm{momentum}} \leftarrow m \cdot \theta_{\mathrm{momentum}}+(1-m) \cdot \theta_{\mathrm{base}}$, where $m \in[0,1)$ is a momentum coefficient. Siamese masking is only used with ViT backbones, where random patches of an image are masked out and only the visible patches are sent through the trainable branch.}
    \label{fig:softcon-main}
\end{figure*}

\Cref{fig:contrast} provides a simplified illustration of the three different contrastive learning designs. Given $N$ raw images, two batches of augmented views are generated and sent through the model to get the corresponding feature vectors $\{z_1,z_2,...,z_N\}$ and $\{z'_1,z'_2,...,z'_N\}$. In the original contrastive learning (\Cref{fig:contrast} a), strict instance discrimination is performed such that the cosine similarity of $z_i$ and its augmented view $z'_i$ is maximized, while the cosine similarities of $z_i$ and all other features $z'_j$ ($i\neq{j}$) are minimized. This formulates an InfoNCE \cite{oord2018representation} loss which identifies one positive candidate from all the samples:

\begin{equation}
\mathcal{L}^{\text {Contrast}}=-\sum_{i=1}^{N} \log \frac{\exp \left({z}_i \cdot {z'_i} / \tau\right)}{\sum_{j=1}^{N} \exp \left({z}_i \cdot {z'_j} / \tau\right)}
\label{equa:1}
\end{equation}

\noindent where $\tau$ is softmax temperature. 
In practice, contrastive learning typically needs a large number of negative samples. Therefore, it inevitably faces the conflict that multiple images besides the augmented view can belong to the same class as the anchor image while the loss strictly pushes them apart.

SupCon tackles this conflict by introducing image labels to generalize contrastive learning to an arbitrary number of
positives. As \Cref{fig:contrast} b shows, images belonging to the same class are pulled together, while images belonging to different classes are pushed away. Formally, SupCon accumulates multiple positive pairs while each item has a sole InfoNCE loss:

\begin{equation}
\mathcal{L}^{\text {SupCon}}=-\sum_{i=1}^{N} \frac{1}{N_p} \sum_{p \in P(i)} \log \frac{\exp \left({z}_i \cdot {z'_p} / \tau\right)}{\sum_{j=1}^N \exp \left({z}_i \cdot {z'_j} / \tau\right)}
\label{equa:2}
\end{equation}

\noindent where $P(i) \equiv\left\{p \in \{1,2,..,N\}: {y_p}={y}_i\right\}$, $y$ is the class label, and $N_p$ is the total number of positive pairs for the anchor feature $z_i$. SupCon successfully alleviates the single-positive dilemma of the original contrastive learning. However, it only works with single-label imagery and the classes are mutually exclusive.

To effectively exploit multi-label annotations, we propose soft contrastive learning as is shown in \Cref{fig:contrast} (c). Samples with more similar label distributions are pulled closer than samples with more dissimilar labels, resulting in a soft positive matching. Specifically, we calculate the pair-wise cosine similarity matrix $Y \in \mathbb{R}^{N \times N}$ of the normalized multi-hot label vectors $\{y_1,y_2,...,y_N\}$ and $\{y'_1,y'_2,...,y'_N\}$. This is done through the dot product of the label vectors: $\mathbf{y} \mathbf{{y'}^T}$. Similarly, we get the similarity matrix $X = \mathbf{z} \mathbf{{z'}^T} \in \mathbb{R}^{N \times N}$ of the feature vectors $\{z_1,z_2,...,z_N\}$ and $\{z'_1,z'_2,...,z'_N\}$. Then, we optimize per-element binary cross entropy loss:

\begin{equation}
\begin{split}
    \mathcal{L}^{\text {SoftCon}}=
    -\sum_{i=1}^{N} \sum_{j=1}^{N} \big(Y_{ij} \cdot \log{\sigma(X_{ij})} \\
    + \left(1-Y_{ij}\right) \cdot \log{\left(1-\sigma(X_{ij})\right)}\big) 
\end{split}
\end{equation}

\noindent where $Y_{ij}$ is a soft score between 0 and 1, and $\sigma(\cdot)$ is the sigmoid function. 

In practice, we follow MoCo-v2 \cite{chen2020improved} and MoCo-v3 \cite{chenempirical} for the implementation of ResNet and ViT backbones, and combine the SoftCon loss with the Contrast loss as:
\begin{equation}
\mathcal{L} = \mathcal{L}^{\text {Contrast}} + \lambda \cdot \mathcal{L}^{\text {SoftCon}}
\label{equa:4}
\end{equation}

\noindent where $\lambda$ is a weighting parameter. This combination is also similarly used by previous works in vision such as \cite{zhu2023multi,bai2022point,wang2022exploring,wang2021dense}. However, our reason is conceptually different: our 9-class multi-label annotations are on a much coarser level than the real world, thus SoftCon alone may restrict the model's capacity to learn complex landscapes. Also, to prevent potential conflict optimization on the same feature embeddings, we use a separate projector for each objective.

\subsection{Continual pretraining with Siamese masking}

Finally, we introduce cross-domain continual pretraining into our framework to boost efficient Earth observation (EO) pretraining. Instead of sophisticated strategies like GFM \cite{mendieta2023towards} that are restricted by RGB input, we revisit the simple weight loading, initializing the backbone with strong vision foundation models. We use DINO \cite{caron2021emerging} weights for ResNet backbones, and DINOv2 \cite{oquab2023dinov2} weights for ViTs. As for the channel difference between natural RGB images and EO multispectral and SAR imagery, we simply let the input embedding layer remain randomly initialized.  Our experiments empirically suggest that though naive, this strategy is both flexible and effective. In addition, to save hardware memory when continually training large
vision Transformers, we adopt Siamese masking inspired by iBOT \cite{zhou2021ibot}, MAE \cite{he2022masked} and MSN \cite{assran2022masked}. Specifically, we randomly mask out a certain percentage of input patches on the trainable branch of
the Siamese contrastive learning framework, and only send the remaining visible patches to the encoder. As the encoder only needs to process a portion of the full patches, both the memory and the training time can be reduced. Note that we do not conduct any reconstruction like MAE or iBOT, but rather view such masking as additional data augmentation. Different from MSN's non-negative online clustering, we verify it also works well in contrastive learning with negative sampling: with a reasonable masking ratio of about 20\%, not only the efficiency, but also the performance can be improved. 
The full pretraining framework of SoftCon is illustrated in \Cref{fig:softcon-main}.


\section{Implementation details}

\subsection{Pretraining} We pretrain SoftCon with ResNet \cite{he2016deep} and ViT \cite{dosovitskiy2020image} backbones on the integrated multi-label dataset SSL4EO-S12-ML. We normalize the 16-bit multispectral images and the 32-bit SAR images to 8-bit with the mean and standard deviation provided in \cite{wang2023ssl4eo}. If there are multiple seasons for one scene, we randomly choose two for the base encoder and the momentum encoder, respectively. Data augmentations follow \cite{wang2023ssl4eo}, including random crop (to the size 224$\times$224), color jitter, greyscale, Gaussian blur, and random flip. 

We adapt MoCo-v2 \cite{chen2020improved} for ResNet50 and MoCo-v3 \cite{chenempirical} for ViT-small and ViT-base, with two separate projectors to get embeddings for the Contrast and the SoftCon loss, respectively. The weighting parameter $\lambda$ in \Cref{equa:2} is 0.1. We set a queue size of 16384 for MoCo-v2, and a batch size of 1024 for both MoCo-v2 and MoCo-v3. The learning rate is warmed-up to 1.5e-4 for 10 epochs followed by cosine decay, and the optimizer is AdamW.

We load ResNet50 weights from DINO\footnote{https://github.com/facebookresearch/dino} \cite{caron2021emerging} and ViT-S/14 and ViT-B/14 weights (without register) from DINOv2\footnote{https://github.com/facebookresearch/dinov2} \cite{oquab2023dinov2}, and conduct continual pretraining for 100 epochs. For ViTs, we randomly mask out 20\% patches and send the remaining ones to the trainable encoder. Training is distributed in two nodes each with 4 NVIDIA A100 GPUs and takes 7-30 hours for different backbones. More details can be seen in the appendix.

\subsection{Downstream tasks}
We evaluate the pretrained backbones by linear probing and fine-tuning in 11 downstream tasks, including 6 land cover land use classification/segmentation datasets: BigEarthNet \cite{sumbul2019bigearthnet}, BigEarthNet-SAR \cite{sumbul2021bigearthnet}, EuroSAT \cite{helber2019eurosat}, EuroSAT-SAR \cite{wang2023feature}, fMoW-sentinel \cite{cong2022satmae} and DFC2020 \cite{schmitt2020ieee}, one change detection dataset OSCD \cite{daudt2018urban}, and 4 multispectral datasets covering different applications from GEO-Bench \cite{lacoste2024geo}: m-so2sat, m-brick-kiln, m-cashew-plantation and m-SA-crop-type. 
\begin{itemize}
    \item BigEarthNet: a large-scale Sentinel-2 multi-label scene classification dataset covering 10 European countries. We use the version of 19 classes, and remove bad patches that are fully covered by seasonal snow or clouds. Following previous works \cite{neumann2019domain,manas2021seasonal,cong2022satmae,wang2023ssl4eo,fuller2024croma}, we train on 1\% or 10\% of the training split (31,166 images), and report micro mean average precision (mAP) on the full validation split (103,944 images).

    \item BigEarthNet-SAR: the paired Sentinel-1 SAR version of the BigEarthNet dataset. We train on 10\% of the training split, and report mAP scores on the full validation split. The splits are aligned with the above multispectral version.

    \item EuroSAT: a 10-class scene classification dataset with 27,000 Sentinel-2 images collected from 34 European countries. Following \cite{fuller2024croma}, we use 16,200 training images and report overall accuracy on 5400 validation images.

    \item EuroSAT-SAR: the paired Sentinel-1 SAR version of the EuroSAT dataset. 

    \item fMoW-sentinel: a large-scale scene classification dataset with 62 classes curated by matching fMoW \cite{christie2018functional} with Sentinel-2 images. Following \cite{fuller2024croma}, we use 10\% of the training split (71,287 images) and report top-1 accuracy on the full validation split (84,939 images).

    \item DFC2020: a 10-class land cover semantic segmentation dataset that was originally collected for 2020 IEEE data fusion contest. We adjust the official test/validation data with 10m resolution labels for 5128 training and 986 testing images and report mean intersection over union (mIoU) scores.

    \item OSCD: a binary change detection dataset consisting of 24 pairs of multispectral images distributed worldwide. We use the official split: 14 training and 10 test pairs.

    \item m-so2sat: a subset of the 17-class So2Sat \cite{zhu2020so2sat} dataset for local climate zone classification. We use the official split from GEO-Bench with 19992/986/986 train/val/test images and report test top-1 accuracies.

    \item m-brick-kiln: a subset of the brick-kiln \cite{lee2021scalable} dataset for binary classification. We use the official split from GEO-Bench with 15063/999/999 train/val/test images and report test accuracies.

    \item m-cashew-plantation: a subset of the cashew-plantation \cite{yin2023mapping} dataset for 7-class semantic segmentation. We use the official split from GEO-Bench with 1350/400/50 train/val/test images and report test mIoU scores.

    \item m-SA-crop-type: a subset of the SA-crop-type \cite{noauthor_source_nodate} dataset for 10-class semantic segmentation. We use the official split from GEO-Bench with 3000/1000/1000 train/val/test images and report test mIoU scores.
    
\end{itemize}

We do a simple grid search for the learning rate for each dataset, using SGD or AdamW optimizer, and train each dataset for 30-100 epochs. We use a common input size 224$\times$224 for all datasets. We freeze the encoder and train a linear layer or the decoder for all GEO-Bench datasets. See the appendix for hyperparameter details.

\section{Results}

\subsection{Land cover classification}

\subsubsection{BigEarthNet and EuroSAT}

\begin{table*}[htbp]
\caption{Linear-probing / fine-tuning results on BigEarthNet-10\% \cite{sumbul2019bigearthnet} and EuroSAT \cite{helber2019eurosat}. We report a comprehensive comparison with SOTA EO foundation models. MS/SAR/RGB represent data modalities. $\dagger$: SoftCon starts from DINO/DINOv2 which were trained on ImageNet/LVD-142M. \# indicates "the number of". *: SkySense employs a mixed architecture (in total 2B parameters) with ViT-L and Geo-context attention to encode MS images; 
875K steps with batch size 240 roughly count to 1000 epochs. Left/right: linear/finetune.
Best scores in \textbf{bold}.}
\label{tab:be-eu}
\centering
\begin{tabular}{llcccccc}
\hline
               & Pretrain dataset & \# Pixels     & Epochs     & Backbone & \# Param. & BE-10\%            & EuroSAT            \\ \hline
Supervised     & -  & -                   & -       & RN50 & 23M     & 83.4               & 98                 \\
MoCo-v2 \cite{chen2020improved,wang2023ssl4eo}        & SSL4EO-S12 (MS) & 70B      & 100       & RN50 & 23M    & 82.1/86.2          & 98.0/99.1          \\
DINO \cite{caron2021emerging,wang2023ssl4eo}           & SSL4EO-S12 (MS) & 70B      & 100       & RN50 & 23M    & 82.0/87.1          & 97.2/99.1          \\
SeCo \cite{manas2021seasonal}           & SeCo (MS)    & 70B         & 100       & RN50 & 23M    & 78.6/82.6          & -/93.1             \\
CACo \cite{mall2023change}           & CACo (MS)  & 70B           & 100       & RN50 & 23M    & -/81.3             & 95.9/-             \\
SoftCon (ours) & {SSL4EO-S12 (MS)}$^\dagger$  & 70B     & 100       & RN50 & 23M    & \textbf{84.8/87.8} & \textbf{98.6/99.3} \\ \hdashline
MoCo-v3 \cite{chenempirical,wang2023ssl4eo}        & SSL4EO-S12 (MS) & 70B      & 100       & ViT-S & 23M   & 82.3/86.1          & 97.7/98.6          \\
DINO \cite{caron2021emerging,wang2023ssl4eo}           & SSL4EO-S12 (MS) & 70B      & 100       & ViT-S & 23M   & 81.7/86.9          & 97.7/99.0          \\
MAE \cite{he2022masked,wang2023ssl4eo}            & SSL4EO-S12 (MS)  & 70B     & 100       & ViT-S & 23M   & 77.5/84.8          & 94.1/98.7          \\
GFM \cite{mendieta2023towards}            & GeoPile (RGB) & 20B        & 100       & Swin-B & 88M  & -/86.3             & -                  \\
SpectralGPT \cite{hong2023spectralgpt}    & fMoW+BigEarthNet (MS) & 12B  & 300       & ViT-B & 86M   & -/87.5             & -/99.2             \\
SatMAE \cite{cong2022satmae}         & fMoW (MS)    & 2.5B         & 200       & ViT-L  
 & 304M  & 80.3/86.2          & 96.6/99.2          \\
CROMA \cite{fuller2024croma}          & SSL4EO-S12 (MS,SAR) & 140B  & 600       & ViT-L & 304M   & 85/88.3            & \textbf{98.0/99.5} \\
SkySense \cite{guo2023skysense}       & SkySense (RGB,MS,SAR) & 97T & 1000$^*$ & ViT-L$^*$  & 517M$^*$  & \textbf{-/88.7}    & -                  \\
SoftCon (ours) & SSL4EO-S12 (MS)$^\dagger$ & 70B      & 100       & ViT-S & 23M   & 85/88.1            & 97.1/99.3          \\
SoftCon (ours) & SSL4EO-S12 (MS)$^\dagger$  & 70B     & 100       & ViT-B & 86M   & \textbf{86.8/88.7} & \textbf{98.0/99.5} \\ \hline
\end{tabular}
\end{table*}

\begin{figure*}[htbp]
    \centering
    \includegraphics[width=0.9\textwidth]{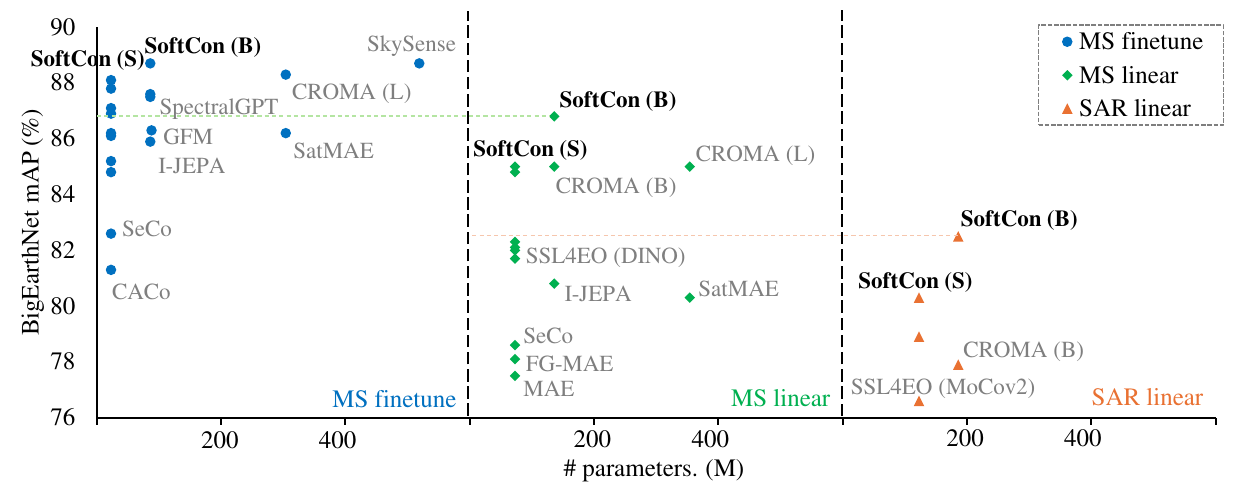}
    \caption{A detailed comparison of transfer learning performances on BigEarthNet-10\%. S/B/L represents ViT-small/base/large. SoftCon (ours) achieves SOTA results with lighter backbones on both linear probing and fine-tuning, and both multispectral and SAR. Our best multispectral linear result is better or comparable to many SOTA models' fine-tuning results; our best SAR result outperforms many multispectral models.}
    \label{fig:scatter-split}
\end{figure*}

We first report SoftCon results on BigEarthNet-10\% \cite{sumbul2019bigearthnet} and EuroSAT \cite{helber2019eurosat} which are most commonly evaluated by SOTA EO foundation models. As can be seen from \Cref{tab:be-eu}, our models outperform most of the existing works in all scenarios. Specifically on BigEarthNet, our ResNet50 improves over other works with the same backbone by a large margin, with 2.7\%/1.6\% increase on linear/fine-tuning compared to the current best model, and 6.2\%/5.2\% increase compared to SeCo \cite{manas2021seasonal}. Our linear probing result for the first time outperforms fully supervised training from scratch. Notably, our small ResNet50 also outperforms many existing large ViT models in both linear and fine-tuning, verifying the effectiveness of ConvNet backbones. 

For ViT backbones, our ViT-small is already comparable to the best existing models like SpectralGPT \cite{hong2023spectralgpt}, CROMA \cite{fuller2024croma} and SkySense \cite{guo2023skysense}. Our ViT-base further pushes forward, achieving a new SOTA of 86.8\% mAP on BE-10\% with linear probing, and the same performance as SkySense in fine-tuning. More specifically, we achieve the same linear probing performance as CROMA \cite{fuller2024croma} with single-modality pretraining (MS v.s. MS+SAR), much fewer model parameters (ViT-S v.s. ViT-L), and much shorter training epochs (100 v.s. 600). Similarly, we reach the fine-tuning performance of SkySense \cite{guo2023skysense} with the above-mentioned advantages, plus much less pretraining data. For better visual comparison, we provide a scatter plot of various SOTA models' transfer learning performances w.r.t model size in  \Cref{fig:scatter-split}.

On EuroSAT \cite{helber2019eurosat}, the best existing models can achieve a top-1 accuracy close to 99\%. This means the dataset is becoming solvable with the fast technological development. As a result, it's less obvious to strictly compare the models' capacities. For example, our ResNet50 is even better than our ViT-base with much fewer parameters. Nevertheless, by comparing both linear and fine-tuning results, our models consistently outperform other works, achieving the same performance as CROMA \cite{fuller2024croma} in both settings.


\subsubsection{BigEarthNet-SAR and EuroSAT-SAR}
\Cref{tab:SAR} reports SoftCon linear probing results in the SAR modality on BigEarthNet-SAR \cite{sumbul2021bigearthnet} and EuroSAT-SAR \cite{wang2023feature} datasets. As we can see, our ResNet50 outperforms MoCo-v2 \cite{chen2020improved} by 2.3\% on BigEarthNet-SAR with 10\% labels, and by 4.7\% on EuroSAT-SAR. Our ViT-S results are much better than MAE \cite{he2022masked} and FG-MAE \cite{wang2023feature} with up to 11.6\% improvement. This verifies again the advantage of contrastive learning over masked image model in producing out-of-the-box representations. Our ViT-B sets a new record of 81.4\% on BigEarthNet-SAR, 3.5\% better than CROMA \cite{fuller2024croma}. \textbf{Notably, this score is already higher than many multispectral models} as compared to \Cref{tab:be-eu}, achieving a great breakthrough as it has always been very difficult for SAR to beat optical in cloud-free scenes. Our results confirm the great potential of SAR foundation models.

\begin{table}[h]
\caption{Linear probing results on BigEarthNet-SAR-10\% \cite{sumbul2021bigearthnet} and EuroSAT-SAR \cite{wang2023feature}.}
\label{tab:SAR}
\centering
\begin{tabular}{lccc}
\hline
        & Backbone & BE-SAR-10\% & EuroSAT-SAR \\ \hline
MoCo-v2 \cite{chen2020improved,wang2023ssl4eo} & RN50     & 76.6       & 82.4        \\
SoftCon (ours) & RN50     & \textbf{78.9}       & \textbf{87.1}        \\ \hdashline
MAE \cite{he2022masked,wang2023ssl4eo}     & ViT-S    & 69.8       & 79.3        \\
FG-MAE \cite{wang2023feature}  & ViT-S    & 71.7       & 80.7        \\
CROMA \cite{fuller2024croma}   & ViT-B    & 77.9       & 87.5        \\
SoftCon (ours) & ViT-S    & 80.3       & 87.1        \\
SoftCon (ours) & ViT-B    & \textbf{82.5}       & \textbf{89.1}        \\ \hline
\end{tabular}
\end{table}

\subsubsection{fMoW-sentinel}

We present linear probing and fine-tuning results on another more difficult land cover classification dataset fMoW-sentinel \cite{cong2022satmae} in \Cref{tab:fmow}. As the table shows, our ViT-small already achieves better performance than all existing models. Our ViT-base pushes further, with 4.8\% better than the current best model CROMA \cite{fuller2024croma} in linear probing, and 1.6\% better in fine-tuning.

\begin{table}[h]
\caption{Linear probing/fine-tuning top-1 accuracy on fMoW-sentinel-10\% \cite{cong2022satmae}. }
\label{tab:fmow}
\centering
\begin{tabular}{llcc}
\hline
               & Backbone & Linear & Finetune        \\ \hline
DINO \cite{caron2021emerging,wang2023ssl4eo}           & ViT-S/16   & 32.6 & 52.8          \\
MAE \cite{he2022masked,wang2023ssl4eo}            & ViT-S/16   & 27.7 & 51.8          \\
SatMAE \cite{cong2022satmae}         & ViT-S/16   & 35.2 & 57.2          \\
I-JEPA \cite{assran2023self}         & ViT-S/16    & 32.4 & 53.5          \\
CROMA \cite{fuller2024croma}          & ViT-L/8     & 39.2 & 59.0          \\
SoftCon (ours) & ViT-S/14   & 39.9 & 59.7          \\
SoftCon (ours) & ViT-B/14   & \textbf{44.0} & \textbf{60.6} \\ \hline
\end{tabular}
\end{table}

\subsection{Land cover segmentation}

\begin{figure*}[]
    \centering
    \includegraphics[width=0.85\linewidth]{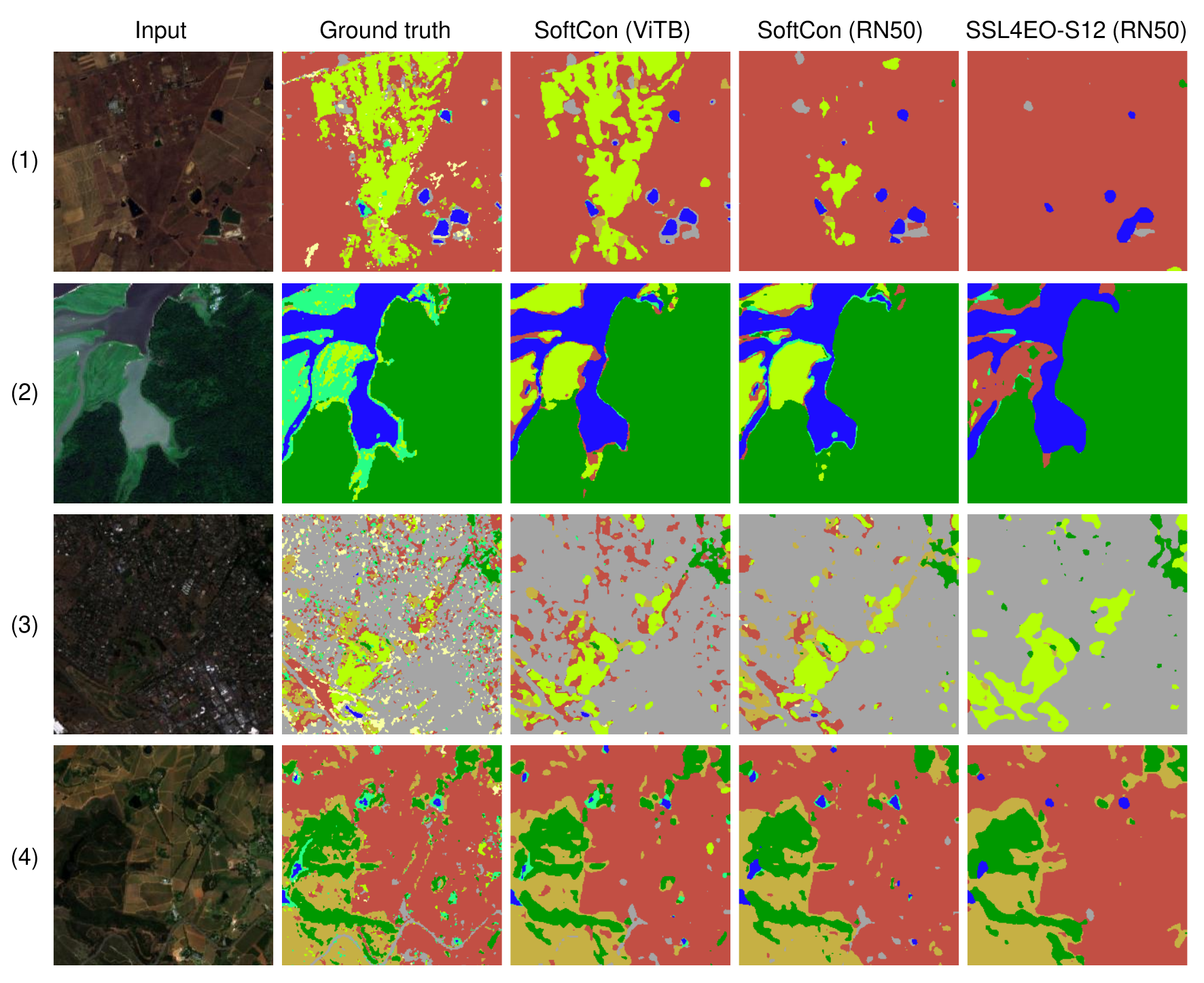}
    \caption{Example segmentation maps on the DFC2020 \cite{schmitt2020ieee} dataset.}
    \label{fig:dfc2020}
\end{figure*}

We report land cover semantic segmentation results on DFC2020 \cite{schmitt2020ieee} as is shown in \Cref{tab:dfc2020}. We fine-tune DeepLabv3+ \cite{chen2018encoder} for ResNet50 and UperNet \cite{xiao2018unified} for ViT backbones. Promisingly, our ResNet50 outperforms MoCo-v2 \cite{chen2020improved} by 3\%, and is also better than existing large ViTs. Our ViT-S outperforms CROMA \cite{fuller2024croma} by 0.9\%, which is further improved by our ViT-base with 0.3\%. For visual comparison, we plot some example segmentation maps in \Cref{fig:dfc2020}. As it shows, SoftCon captures more accurate and more fine-grained semantic information compared to SSL4EO-S12 \cite{wang2023ssl4eo}.

\begin{table}[h]
\caption{Fine-tuning mIoU results on DFC2020 \cite{schmitt2020ieee}.}
\label{tab:dfc2020}
\centering
\begin{tabular}{llc}
\hline
                        & Backbone & mIoU \\ \hline
Supervised              & RN50     & 42.9    \\
MoCo-v2 \cite{chen2020improved,wang2023ssl4eo}                 & RN50     & 47.3    \\
\textbf{SoftCon (ours)} & RN50     & \textbf{50.3}    \\ \hdashline
MAE \cite{he2022masked,wang2023ssl4eo}                     & ViT-S    & 48.0    \\
SatMAE \cite{cong2022satmae}                  & ViT-L    & 44.1    \\
CROMA \cite{fuller2024croma}                   & ViT-L    & 49.8    \\
\textbf{SoftCon (ours)} & ViT-S    & 50.7    \\
\textbf{SoftCon (ours)} & ViT-B    & \textbf{51.0}    \\ \hline
\end{tabular}
\end{table}

\subsection{Change detection}

We report binary change detection results on OSCD \cite{daudt2018urban} as is shown in Table \Cref{tab:oscd}. We freeze the backbone and fine-tune a simple U-Net \cite{ronneberger2015u} for segmentation. The differences in feature maps between two timestamps are input to the network. SoftCon outperforms SeCo and SSL4EO in both recall and F1-score. The low precision score is due to the significant class unbalance, i.e.: predicting all pixels as unchanged would result in a good precision score. The combined F1-score highlights the superior performance of SoftCon.

\begin{table}[h]
\caption{Results with frozen encoders on OSCD \cite{daudt2018urban}. }
\label{tab:oscd}
\centering
\begin{tabular}{lcccc}
\toprule
               & Backbone & Precision     & Recall        & F1            \\ \hline
Rand. Init.    & RN50     & 72.3          & 13.8          & 23.1          \\
SeCo \cite{manas2021seasonal}   & RN50     & \textbf{74.9} & 17.5          & 28.3          \\
SSL4EO \cite{wang2023ssl4eo} & RN50     & 70.2          & 23.4          & 35.1          \\
SoftCon (ours) & RN50     & 66.6          & \textbf{29.2} & \textbf{40.6} \\ \bottomrule
\end{tabular}
\end{table}

\subsection{Other domain-specific applications}

\begin{table*}[]
\caption{Transfer results with frozen encoders on four Sentinel-2 tasks in GEO-Bench \cite{lacoste2024geo}. We report top-1 accuracy/mIoU for classification/segmentation, respectively.}
\label{tab:geobench}
\centering
\begin{tabular}{llcccc}
\hline
               &       & m-so2sat      & m-brick-kiln  & m-cashew-plantation & m-SA-crop-type \\ \hline
GEO-Bench (FT) & RN50  & 52.8          & 98.7          & 44.7                & 29.9           \\ \hdashline
CROMA \cite{fuller2024croma}          & ViT-B & 49.2          & 91.0          & -                   & 31.4           \\
OFA-Net \cite{xiong2024one}        & ViT-B & 46.0          & 91.3          & 37.4                & \textbf{32.0}  \\
SoftCon (ours) & ViT-S & 49.9          & 92.6          & 44.4                & 31.5           \\
SoftCon (ours) & ViT-B & \textbf{52.0} & \textbf{95.2} & \textbf{49.6}       & 31.5           \\ \hline
\end{tabular}
\end{table*}

Finally, we evaluate SoftCon on four other domain-specific applications with the corresponding Sentinel-2 dataset collections from GEO-Bench \cite{lacoste2024geo}. These include two classification datasets m-so2sat and m-brick-kiln, and two segmentation datasets m-cashew-plantation and m-SA-crop-type. We conduct frozen-encoder training following \cite{xiong2024one}, with a linear classifier for classification tasks and UperNet decoder for segmentation tasks. The results are shown in \Cref{tab:geobench}. We compare SoftCon with CROMA \cite{fuller2024croma} and a recent work OFA-Net \cite{xiong2024one}, and report the official fine-tuning (from timm \cite{rw2019timm}) results in \cite{lacoste2024geo} for reference. The table shows that SoftCon significantly outperforms both CROMA and OFA-Net on three tasks, while only slightly worse than OFA-Net on m-SA-crop-type. Notably, our results with frozen-encoder outperform the official results with full fine-tuning on the two segmentation datasets, and only slightly worse on m-so2sat.

\section{Ablation and discussion}

For all ablation studies, we conduct linear probing experiments on BigEarthNet \cite{sumbul2019bigearthnet}.

\subsubsection{SoftCon loss}
We ablate ResNet50 results on the soft contrastive loss in \Cref{equa:4} which are shown in \Cref{tab:softcon}. In line with our explanation in \Cref{subsec:softcon}, SoftCon alone is worse than Contrast alone since the 10 Dynamic World classes are too coarse-grained to fully represent the real-world semantics. However, combining the two losses provides significant benefits. When we degrade the SoftCon to SupCon in \Cref{equa:2}, the performance drops as expected, while still better than contrastive loss alone. These results verify the effectiveness of using existing free annotations to boost Earth observation pretraining. Additionally, we ablate the weighting parameter $\lambda$ in \Cref{equa:4} and suggest a best value of 0.1. Note all results in this ablation are without continual pretraining, which we will ablate next.

\begin{table*}[ht]
\caption{Ablation study on the SoftCon loss and the weighting parameter.}
\label{tab:softcon}
\centering
\begin{tabular}{lcc}
\hline
                   & BE-1\% & BE-10\% \\ \hline
Contrast only      & 78.9   & 82.1    \\
SoftCon only       & 76.0   & 80.4   \\
Contrast + SupCon  & 79.3   & 83.0    \\
Contrast + SoftCon & \textbf{79.8}   & \textbf{83.6}   \\ 
\hline
\end{tabular}
\hspace{2em}
\begin{tabular}{lcc}
\hline
             & BE-1\%        & BE-10\%       \\ \hline
$\lambda=0.01$         & 79.3          & 82.9          \\
\textbf{$\lambda=0.1$} & \textbf{79.8} & \textbf{83.6} \\
$\lambda=0.5$          & 79.6          & 83.4          \\
$\lambda=1.0$          & 79.5          & 83.3  
    \\ \hline
\end{tabular}

\end{table*}

\subsubsection{Continual pretraining}

We ablate the continual pretraining strategy in \Cref{tab:continual}, which verifies the benefits of loading vision foundation models instead of pretraining from scratch. Also, we compare the performance of different vision models from ImageNet supervised weights to modern self-supervised weights. Interestingly, the stronger the vision model, the continual pretraining performance is also better. We report ImageNet top-1 linear scores of the corresponding models as a reference, which are well aligned with the cross-domain continual pretraining results. 

In addition, we empirically find parameter-efficient fine-tuning (PEFT) techniques such as BitFit \cite{zaken2021bitfit}, prompt tuning \cite{jia2022visual} and LoRA \cite{hu2021lora}, can be used for parameter-efficient continual pretraining, while not yet reaching a close performance as continually training all parameters. We report such preliminary results in the appendix.

\begin{table}[]
\caption{Ablation study on the source models of continual pretraining. \textit{cont.} indicates continual pretraining.}
\label{tab:continual}
\centering
\begin{tabular}{llccc}
\hline
                 & Backbone & BE-1\%        & BE-10\%       & {\color[HTML]{9B9B9B} ImageNet}      \\ \hline
w/o cont.        & RN50     & 79.8          & 83.6          & {\color[HTML]{9B9B9B}-}             \\
cont. (ImageNet) & RN50     & 80.2          & 83.9          & {\color[HTML]{9B9B9B}-}             \\
cont. (MoCov3 \cite{chenempirical})   & RN50     & 80.9          & 84.2          & {\color[HTML]{9B9B9B}74.6}          \\
cont. (DINO \cite{caron2021emerging})     & RN50     & \textbf{81.4}          & \textbf{84.8}          & {\color[HTML]{9B9B9B}75.3}          \\ \hdashline
cont. (DINOv2 \cite{oquab2023dinov2})   & ViT-S    & \textbf{82.6} & \textbf{85.0} & {\color[HTML]{9B9B9B}81.1} \\ \hline
\end{tabular}

\end{table}

\subsubsection{Siamese masking}

Finally, we ablate the Siamese masking strategy introduced mainly to boost training efficiency. As can be seen from \Cref{tab:mask}, masking 50\% of the patches can save almost half of the GPU memory without much performance drop. Interestingly, masking 20\% of the patches can lead to even better results than no masking. This suggests that MAE-like random masking can be seen as an effective data augmentation strategy in contrastive learning. From another perspective, this design implicitly introduces the idea of masked image modeling that the model needs to know what information the masked patches have according to the unmasked branch, after which it knows the encoded features are similar. This could potentially also be one reason why the ideal masking ratio is much smaller compared to MAE, as the implicit masked image modeling plus the contrastive learning is a more challenging optimization task.

\vspace{0.5em}
\noindent In summary, \Cref{tab:gain} provides an overview of the performance gains of each of our proposed components. It is worth noting that when introducing the multi-label annotation, the supervised contrastive learning method is implicitly introduced, which can be improved by our proposed soft contrastive learning as shown by the ablation in \Cref{tab:softcon}. Furthermore, when using the continual pretraining method, the vision dataset (e.g. ImageNet) is implicitly used for the vision model weights.

\begin{table}[]
\caption{Ablation study on the Siamese masking ratio.}
\label{tab:mask}
\centering
\begin{tabular}{lccc}
\hline
               & GPU memory & BE-1\%        & BE-10\%       \\ \hline
w/o masking    & 43G        & 82.3          & 84.5          \\ 
masking (20\%) & 36G        & \textbf{82.6} & \textbf{85.0} \\
masking (50\%) & 25G        & 81.7          & 84.4 \\ \hline        
\end{tabular}
\end{table}

\begin{table}[ht]
\caption{Performance gains of different components under linear probing on BigEarthNet-10\%. The explicitly introduced components are highlighted in \textbf{bold}.}
\label{tab:gain}
\centering
\begin{tabular}{cccc}
\hline
Method             & Backbone & Dataset                & Performance gain \\ \hline
Contrast           & RN50 & SSL4EO-S12             & -    \\ \hdashline
(+ SupCon)         & RN50 & \textbf{+ multi-label} & +0.9 \\
\textbf{+ SoftCon} & RN50 & -                      & +0.6 \\
\textbf{+ cont.}   & RN50 & (+ ImageNet)           & +1.2 \\
\textbf{+ mask}    & ViT-S & -                      & +0.5 \\ \hline
\end{tabular}
\end{table}

\section{Conclusion}
This work revisits two important free resources beyond pure satellite imagery for efficient Earth observation pretraining: open land cover products and open vision foundation models. We build a large-scale multi-label dataset by matching an unsupervised pretraining dataset SSL4EO-S12 with Dynamic World land cover maps. To effectively utilize the multi-label annotations, we propose a novel soft contrastive learning method that allows soft matching between images with different label distributions. Meanwhile, we introduce strong vision models such as DINO and DINOv2 to a simple but flexible continual pretraining framework with Siamese masking. We efficiently train multispectral and SAR foundation models of both CNN and Transformer backbones that achieve SOTA performances in 10 out of 11 downstream tasks. 

There are two main limitations of this work. First, our models target mainly medium-resolution data, while high-resolution images with more fine-grained semantic information remain to be explored. Second, retraining vision foundation models without any constraints inevitably makes the model forget knowledge from the source vision domain. Future work will explore more effective and flexible multimodal continual pretraining methods to build strong cross-domain cross-modal foundation models.

\section*{Acknowledgments}
This work is jointly supported by the European Commission through the project “ThinkingEarth—Copernicus Foundation Models for a Thinking Earth” under the Horizon 2020 Research and Innovation program (Grant Agreement No. 101130544), by the Helmholtz Association through the
Framework of HelmholtzAI, grant ID: ZT-I-PF-5-01 – Local Unit Munich Unit @Aeronautics,
Space and Transport (MASTr), by the European Commission through the project "EvoLand" under the Horizon 2020 Research and Innovation program (Grant Agreement No. 101082130), by the Excellence Strategy of the Federal Government and the Lander through the TUM Innovation Network EarthCare, by the German Federal Ministry of Education and Research (BMBF) in the framework of the international future AI lab “AI4EO – Artificial Intelligence for Earth Observation: Reasoning, Uncertainties, Ethics and Beyond” (grant number: 01DD20001) and by Munich Center for Machine Learning. The compute is supported by the Helmholtz Association’s Initiative and Networking Fund on the HAICORE@FZJ partition.



 

\bibliographystyle{IEEEtran}
\bibliography{refs}

\newpage

\appendices

\section{SSL4EO-S12-ML Dataset}

\subsection{General information}

SSL4EO-S12-ML dataset is a large-scale multi-label land cover land use classification dataset. It consists of 780,371 multispectral Sentinel-2 images with size 264$\times$264, divided into 247,377 non-overlapping scenes each with 1-4 multi-seasonal patches. Each image has a multi-label annotation from one or more categories in 9 land cover land use classes. 

\subsubsection{Data source}
The Sentinel-2 images are from SSL4EO-S12 \cite{wang2023ssl4eo}, a multi-modal multi-temporal dataset specifically designed for large-scale self-supervised learning. The dataset consists of 4-seasonal Sentinel-1/2 SAR-optical images from 251,079 non-overlapping locations in the world, covering a wide range of geographical and temporal diversities. The multi-label annotations are derived from Dynamic World \cite{brown2022dynamic}, a near-real-time dataset that provides continuous pixel-level land cover monitoring in 9 semantic classes. The Dynamic World segmentation maps are automatically generated by algorithms developed on high-quality training data. We integrate the noisy maps into rather accurate scene-level classification labels.

\subsubsection{Dataset curation}
Based on the geospatial coordinates and acquisition time, we match each SSL4EO-S12 L1C multispectral image with its corresponding Dynamic World land cover map in Google Earth Engine. Due to the effect of clouds, a few images do not have a corresponding segmentation map, which are then dropped in our finall dataset. In total, there are 247,377 scenes with a successful match for at least one season. Then, we gather the pixel labels into scene-level multi-labels for each image, resulting in 780,371 labeled individual images. The workflow has ben shown in Figure 2.


\subsection{Dataset statistics}

\subsubsection{Label number distribution}
\begin{figure}[h]
    \centering
    \includegraphics[width=\columnwidth]{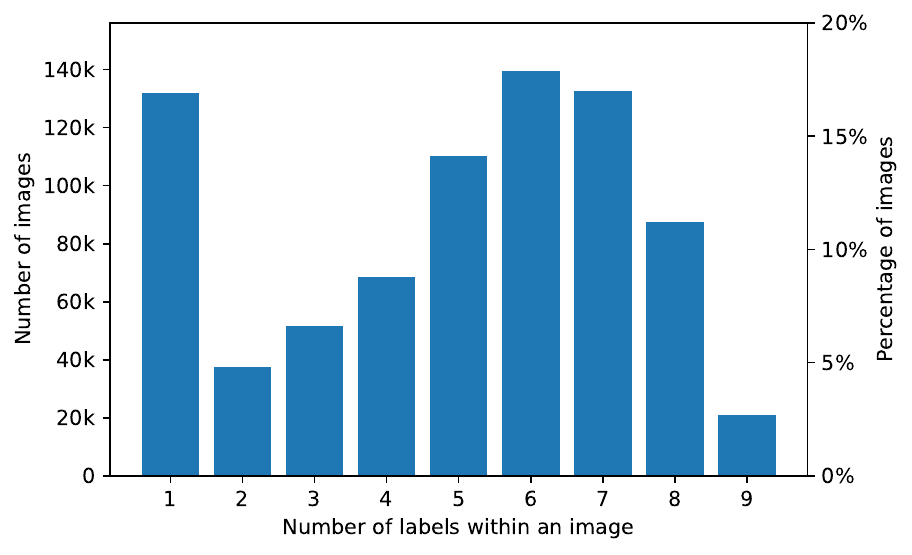}
    \caption{The distribution of label numbers within each image.}
    \label{fig:dist_label}
\end{figure}

\Cref{fig:dist_label} shows the distribution of the number of labels within each image. About 17\% images contain a single label, while 70\% images contain 4 or more labels.

\subsubsection{Class distribution}
\begin{figure}[h]
    \centering
    \includegraphics[width=\columnwidth]{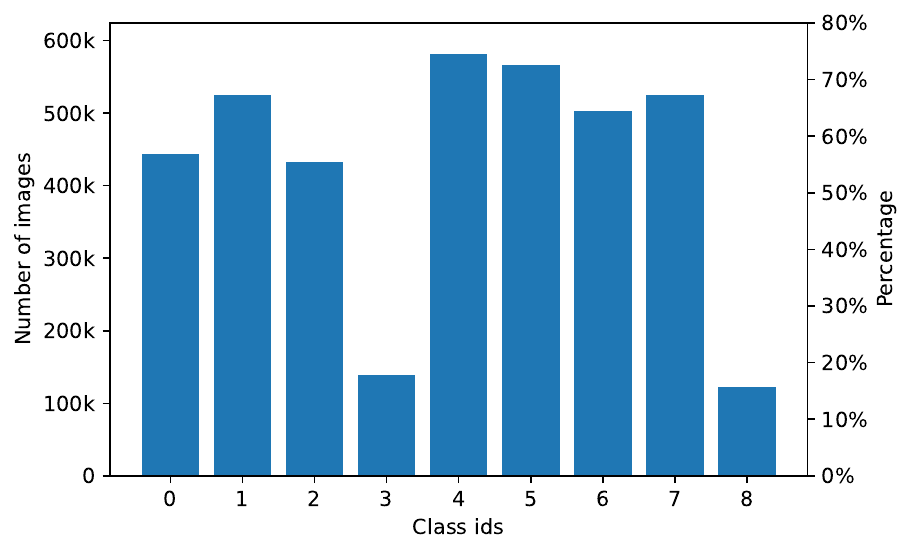}
    \caption{The distribution of image numbers for different classes.}
    \label{fig:dist_cls}
\end{figure}

\Cref{fig:dist_cls} shows the distribution of the number of images for each class. The number of images is well balanced in 7 common classes, while flooded vegetation and snow/ice are less represented.

\subsubsection{Season distribution}
\begin{figure}[h]
    \centering
    \includegraphics[width=\columnwidth]{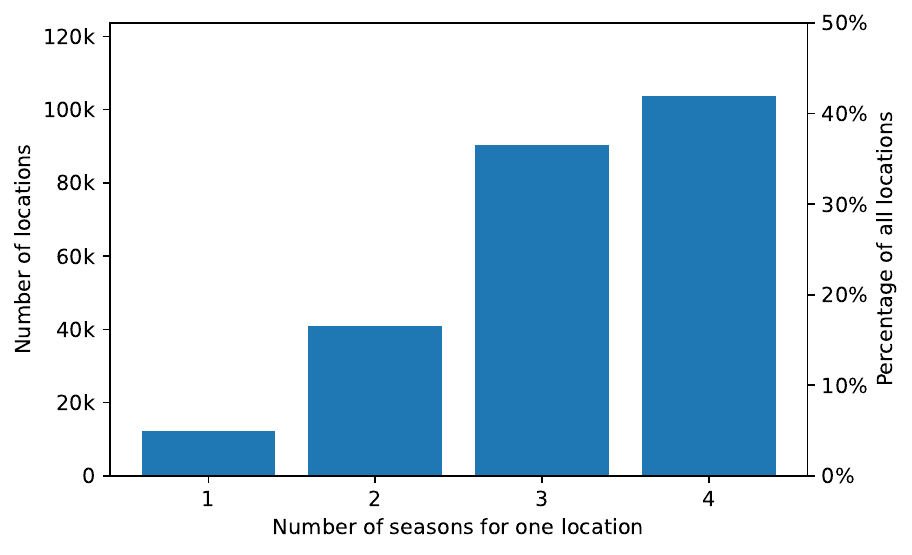}
    \caption{The distribution of season numbers for each location.}
    \label{fig:dist_season}
\end{figure}

\Cref{fig:dist_season} shows the distribution of the number of seasonal patches for each location. More than 40\% locations have all 4 seasons, and more than 95\% locations have at least 2 seasons.

\subsection{Benchmark}

We provide a preliminary benchmark on SSL4EO-S12-ML as a supervised multi-label classification task in \Cref{tab:benchmark}. We split the dataset into 80\% training data and 20\% testing data according to the non-overlapping locations, and report micro and macro mAP on the test split. We use 10\% of the training data and the full testing data for the benchmark. The table shows that this multi-label classification task is rather easy to solve, with a supervised micro mAP reaching 98.2\%. This is similar to the training metric, indicating the balanced dataset split and rather good label quality. In addition, we also test the dataset as a downstream task to evaluate pretrained models. As is shown in the table, consistent performances as other popular datasets in the main paper are observed: 1) pretrained models improve upon random initialization; 2) SoftCon outperforms existing models such as SSL4EO. Therefore, SSL4EO-S12-ML can be considered as a global multi-label classification benchmark dataset that complements existing datasets such as BigEarthNet \cite{sumbul2019bigearthnet} which covers only Europe.

\begin{table}[]
\caption{Benchmark results of SSL4EO-S12-ML dataset. We use 10\% of training data.}
\label{tab:benchmark}
\centering
\begin{tabular}{lcc}
\hline
                    & mAP (micro) & mAP (macro) \\ \hline
rand. init.         & 93.7        & 84.4        \\
supervised          & 98.2        & 94.7        \\ \hdashline
SSL4EO (linear) \cite{wang2023ssl4eo}     & 95.8        & 88.7        \\
SoftCon (linear)    & 96.5        & 90.2        \\
SoftCon (fine-tune) & 98.8        & 96.1       \\ \hline
\end{tabular}
\end{table}

\subsection{Data examples}

\begin{figure*}
    \centering
    \begin{subfigure}[]{0.3\textwidth}
        \centering
        \includegraphics[width=\textwidth]{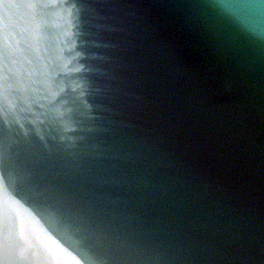}
        \caption{water \vspace{1em}}
    \end{subfigure}
    \begin{subfigure}[]{0.3\textwidth}
        \centering
        \includegraphics[width=\textwidth]{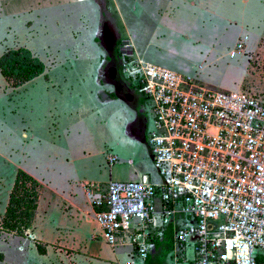}
        \caption{water, trees, grass, flooded vegetation, crops, shrub and scrub, built}
    \end{subfigure}    
    \begin{subfigure}[]{0.3\textwidth}
        \centering
        \includegraphics[width=\textwidth]{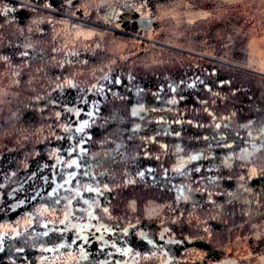}
        \caption{crops, shrub and scrub, bare \vspace{1em}}
    \end{subfigure}
    \begin{subfigure}[]{0.3\textwidth}
        \centering
        \vspace{1em}
        \includegraphics[width=\textwidth]{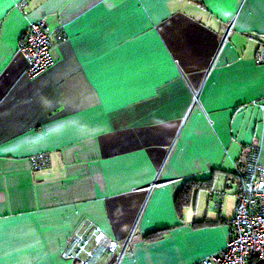}
        \caption{water, trees, grass, crops, shrub and scrub, built, bare}
    \end{subfigure}
    \begin{subfigure}[]{0.3\textwidth}
        \centering
        \includegraphics[width=\textwidth]{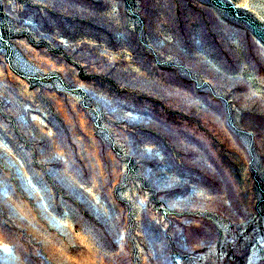}
        \caption{grass, crops, shrub and scrub, bare}
    \end{subfigure}    
    \begin{subfigure}[]{0.3\textwidth}
        \centering
        \includegraphics[width=\textwidth]{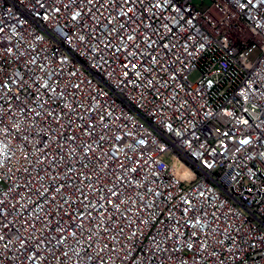}
        \caption{trees, grass, shrub and scrub, built, bare }
    \end{subfigure}

    \caption{SSL4EO-S12-ML example images with multi-label annotations.}
    \label{fig:example}
\end{figure*}

\Cref{fig:example} provides some example images and corresponding multi-labels in the SSL4EO-S12-ML dataset covering different landscapes.

\section{Parameter Efficient Continual Pretraining}

Besides SoftCon weight initialization, we also explored other continual pretraining strategies, such as the adaptation of parameter efficient fine tuning (PEFT) techniques. PEFT is usually used in fine-tuning foundation models to specific downstream tasks with low cost. We studied the feasibility of adapting it to continual pretraining, termed parameter efficient continual pretraining (PECP), to build foundation models for a target domain such as Earth observation. 

We study PECP with three representative PEFT techniques: bias-tuning like BiTFiT \cite{zaken2021bitfit}, visual prompt tuning \cite{jia2022visual}, and low-rank adapter (LoRA) \cite{hu2021lora}. To fully evaluate the capacity of large foundation models, we transfer DINOv2 ViT-Large with about 300M parameters which is on par with commonly studied models in PEFT. To save computing costs and to rule out other factors, we use a simple MAE \cite{he2022masked} for pretraining.

\begin{table*}[]
\caption{Linear probing mAP scores on BigEarthNet-10\% with different pretraining strategies. We report pretraining with full SSL4EO-S12 data and 10\% data. PEFT indicates directly transferring the DINOv2 model on BigEarthNet with PEFT techniques. *: SoftCon represents our SoftCon-style fully continual pretraining used in the main paper.}
\label{tab:pecp}
\centering
\begin{tabular}{llccc}
\hline
                      & pretrain module      & \# pretrain params. & pretrain 100\% & pretrain 10\%  \\ \hline
rand. init.           & -                    & -                  & 64.4          & -             \\
supervised            & -                    & -                  & 74.7          & -             \\
PEFT (bias)           & -                    & -                  & 69.2          & -             \\
PEFT (lora)           & -                    & -                  & 69.2          & -             \\ \hdashline
pretrain from scratch & all                  & 305M               & 79.2          & 72.8          \\
full CP (SoftCon*)     & all                  & 305M               & \textbf{81.0} & \textbf{78.1} \\
PECP (base)           & patch embed          & 2.6M               & 70.6          & 69.1          \\
PECP (bias)           & patch embed + bias   & 2.9M               & 72.7          & 70.2          \\
PECP (prompt)         & patch embed + prompt & 3.0M               & 74.0          & 72.1          \\
PECP (lora)           & patch embed + lora   & 5.8M               & 74.2          & 73.9          \\ \hline
\end{tabular}
\end{table*}

\Cref{tab:pecp} presents linear probing results on BigEarthNet with different pretraining strategies. Firstly, as the table shows, straightforward PEFT from DINOv2 does provide benefits compared to random frozen encoder. However, such benefits are rather limited compared to in-domain pretraining. Secondly, compared to pretraining from scratch, SoftCon-style continual pretraining offers significant improvement, especially when the in-domain data is limited in size: continual pretraining on 10\% data is comparable to pretraining full data from scratch. Thirdly, all PECP strategies provide reasonable benefits with only a small fraction of parameters trainable, outperforming direct PEFT. Among them, LoRA performs slighter better than others. Similar to full continual pretraining, PECP also remains the performance when the size of pretraining data is restricted. Lastly, all the simple PECP strategies are not close to full continual pretraining, restricting the practical usage. This preliminary study motivates us to use full continual pretraining for SoftCon in the main papaer. However, with more research towards advanced designs, we believe PECP bears potential for future work, as the flexible adapters can pave the way towards unified cross-domain foundation models.


\section{Implementation Details}
\subsection{Pretraining}

We pretrain SoftCon with ResNet \cite{he2016deep} and ViT \cite{dosovitskiy2020image} backbones on the proposed multi-label dataset SSL4EO-S12-ML, which consists of 247,377 scenes with 1-4 seasons. We normalize the 16-bit images to 8-bit with the mean and standard deviation provided in \cite{wang2023ssl4eo}. If there are multiple seasons for one scene, we randomly choose two for the base encoder and the momentum encoder, respectively. Data augmentations follow \cite{wang2023ssl4eo}, including random crop (to the size 224$\times$224), color jitter, greyscale, Gaussian blur, and random flip. 

We adapt MoCo-v2 \cite{chen2020improved} for ResNet50 and MoCo-v3 \cite{chenempirical} for ViT-S/14 and ViT-B/14, with two separate projectors to get embeddings for the Contrast and the SoftCon loss, respectively. Each projector consists of two linear layers and one ReLU activation function in between.
The weighting parameter $\lambda$ trading off the two losses is 0.1. We set a queue size of 16384 with a batch size of 256 for MoCo-v2, and a batch size of 1024 for MoCo-v3. For MoCo-v2, the learning rate starts from 0.03, followed by cosine decay to 0; for MoCo-v3, the learning rate is warmed-up to 1.5e-4 for 10 epochs, followed by cosine decay to 0. The optimizer is SGD for MoCo-v2 and AdamW for MoCo-v3. The softmax temperature is 0.2 for both MoCo-v2 and MoCo-v3.

We load ResNet50 weights from DINO \cite{caron2021emerging} and ViT-S/14 and ViT-B/14 weights (without register) from DINOv2 \cite{oquab2023dinov2}, and conduct continual pretraining for 100 epochs. For ViTs, we randomly mask out 20\% patches and send the remaining patches to the trainable encoder. All patches without masking are sent to the momentum encoder. Training is distributed in two nodes each with 4 NVIDIA A100 GPUs. Detailed compute and training time for different backbones are shown in \Cref{tab:compute}.

\begin{table}[h]
    \caption{Compute and pretraining time.}
    \label{tab:compute}
    \centering
\begin{tabular}{clcc}
\hline
Modality             & Backbone & GPUs   & Training time \\ \hline
\multirow{3}{*}{MS}  & RN50     & 4xA100 & 21h           \\
                     & ViT-S/14 & 4xA100 & 25h           \\
                     & ViT-B/14 & 8xA100 & 15h           \\ \hdashline
\multirow{3}{*}{SAR} & RN50     & 4xA100 & 7h            \\
                     & ViT-S/14 & 4xA100 & 8h            \\
                     & ViT-B/14 & 8xA100 & 7h            \\ \hline
\end{tabular}

\end{table}

\subsection{Downstream tasks}

\subsubsection{Downstream tasks}
We evaluate the pretrained backbones by linear probing and fine-tuning in 8 downstream tasks, including 4 land cover land use classification/segmentation datasets: BigEarthNet \cite{sumbul2019bigearthnet}, EuroSAT \cite{helber2019eurosat}, fMoW-sentinel \cite{cong2022satmae} and DFC2020 \cite{schmitt2020ieee}, and 4 multispectral datasets covering different applications from GEO-Bench \cite{lacoste2024geo}: m-so2sat, m-brick-kiln, m-cashew-plantation and m-SA-crop-type. 

For classification datasets, we conduct linear probing and fine-tuning; for DFC2020, we fine-tune DeepLabv3+ \cite{chen2018encoder} for ResNet backbone and UperNet \cite{xiao2018unified} for ViT backbones; for GEO-Bench segmentation datasets, we freeze the encoder and train a UperNet decoder. For linear probing with ViT backbones, we pick either the output features from the last block, or the concatenation of features from the last 4 blocks following DINOv2 \cite{oquab2023dinov2}. We do a simple grid search for the learning rate for each dataset. Specific hyperparameters for each dataset is summarized in the following tables.

\begin{table}[h]
\caption{DFC2020 fine-tuning hyperparameters.}
\label{tab:hyper-dfc2020}
\centering
\begin{tabular}{lcc}
\hline
              & DFC2020 (ResNet)                                                                                  & DFC2020 (ViT)                                                                                 \\ \hline
Backbone      & ResNet50                   & ViT-B/14, ViT-S/14 \\
Input size    & 224x224                                                                                         & 224x224                                                                                         \\
Augmentation  & \begin{tabular}[c]{@{}c@{}}ResizedCrop (0.5,2.0),\\ HorizontalFlip,\\ VerticalFlip\end{tabular} & \begin{tabular}[c]{@{}c@{}}ResizedCrop (0.5,2.0),\\ HorizontalFlip,\\ VerticalFlip\end{tabular} \\
Batch size    & 16                                                                                              & 16                                                                                              \\
Learning rate & 1e-3                                                                                        & 5e-3                                                                                        \\
LR schedule   & poly                                                                                            & poly                                                                                            \\
Optimizer     & SGD                                                                                             & AdamW                                                                                           \\
Weight decay  & 5.00E-04                                                                                        & 5.00E-02                                                                                        \\
Warm-up       & 0                                                                                               & 1000                                                                                            \\
Iter         & 20K                                                                                             & 40K                                                                                             \\
Head          & DeepLabv3+                                                                                      & UperNet                                                                                         \\ \hline
\end{tabular}
\end{table}

\begin{table*}[]
    \caption{Linear probing hyperparameters on BigEarthNet, EuroSAT and fMoW-S2.}
    \label{tab:cls_lc}
    \centering
\begin{tabular}{lccc}
\hline
              & BigEarthNet                                                                      & EuroSAT                                                                         & fMoW-S2                                                                         \\ \hline
Backbone     & ViT-B/14, ViT-S/14     & ViT-B/14, ViT-S/14   & ViT-B/14, ViT-S/14 \\
Input size    & 224x224                                                                          & 224x224                                                                         & 224x224                                                                         \\
Augmentation  & \begin{tabular}[c]{@{}c@{}}ResizedCrop (0.8,1.0), \\ HorizontalFlip\end{tabular} & \begin{tabular}[c]{@{}c@{}}ResizedCrop (0.2,1.0),\\ HorizontalFlip\end{tabular} & \begin{tabular}[c]{@{}c@{}}ResizedCrop (0.2,1.0),\\ HorizontalFlip\end{tabular} \\
Batch size    & 256                                                                              & 256                                                                             & 1024                                                                            \\
Learning rate & 0.1                                                                              & 1e-3                                                                           & 4e-4                                                                        \\
LR schedule   & cos                                                                              & step                                                                            & cos                                                                             \\
Optimizer     & SGD                                                                              & SGD                                                                             & AdamW                                                                           \\
Weight decay  & 0                                                                                & 0                                                                               & 0.01                                                                            \\
Warm-up       & 0                                                                                & 0                                                                               & 0                                                                               \\
Epoch        & 100                                                                              & 100                                                                             & 100                                                                             \\
Feature block & 4          & 1          & 4           \\ \hline
\end{tabular}

\end{table*}

\begin{table*}[]
\caption{Fine-tuning hyperparameters on BigEarthNet, EuroSAT and fMoW-S2.}
\label{tab:cls_ft}
\centering
\begin{tabular}{lccc}
\hline
              & BigEarthNet                                                                                     & EuroSAT                                                                         & fMoW-S2                                                                                         \\ \hline
Backbone     & ViT-B/14, ViT-S/14     & ViT-B/14, ViT-S/14   & ViT-B/14, ViT-S/14 \\
Input size    & 224x224                                                                                         & 224x224                                                                         & 224x224                                                                                         \\
Augmentation  & \begin{tabular}[c]{@{}c@{}}ResizedCrop (0.8,1.0),\\ HorizontalFlip,\\ VerticalFlip\end{tabular} & \begin{tabular}[c]{@{}c@{}}ResizedCrop (0.2,1.0),\\ HorizontalFlip\end{tabular} & \begin{tabular}[c]{@{}c@{}}ResizedCrop (0.2,1.0),\\ HorizontalFlip,\\ mixup\&cutmix\end{tabular} \\
Batch size    & 256                                                                                             & 256                                                                             & 1024                                                                                            \\
Learning rate & 1e-4                                                                                        & 1e-3                                                                        & 4e-4                                                                                        \\
LR schedule   & cos                                                                                             & step                                                                            & cos                                                                                             \\
Optimizer     & AdamW                                                                                           & SGD                                                                             & AdamW                                                                                           \\
Weight decay  & 0.01                                                                                            & 0                                                                               & 0.05                                                                                            \\
Warm-up       & 0                                                                                               & 0                                                                               & 0                                                                                               \\
Epoch        & 50                                                                                              & 100                                                                             & 50                                                                                              \\ \hline
\end{tabular}
\end{table*}

\begin{table*}[]
    \caption{ResNet linear probing (left) and fine-tuning (right) hyperparameters on BigEarthNet and EuroSAT.}
    \label{tab:r50_cls_lc}
    \centering
\begin{tabular}{lcc}
\hline
              & BigEarthNet                                                                      & EuroSAT                                                                         \\ \hline
Backbone      & RN50                                                                             & RN50                                                                            \\
Input size    & 224x224                                                                          & 224x224                                                                         \\
Augmentation  & \begin{tabular}[c]{@{}c@{}}ResizedCrop (0.8,1.0), \\ HorizontalFlip\end{tabular} & \begin{tabular}[c]{@{}c@{}}ResizedCrop (0.2,1.0),\\ HorizontalFlip\end{tabular} \\
Batch size    & 256                                                                              & 256                                                                             \\
Learning rate & 8                                                                                & 8                                                                               \\
LR schedule   & step                                                                             & step                                                                            \\
Optimizer     & SGD                                                                              & SGD                                                                             \\
Weight decay  & 0                                                                                & 0                                                                               \\
Warm-up       & 0                                                                                & 0                                                                               \\
Epoch         & 100                                                                              & 100                                                                             \\ \hline
\end{tabular}
\begin{tabular}{lcc}
\hline
              & BigEarthNet                                                                                       & EuroSAT                                                                         \\ \hline
Backbone      & RN50                                                                                              & RN50                                                                            \\
Input size    & 224x224                                                                                           & 224x224                                                                         \\
Augmentation  & \begin{tabular}[c]{@{}c@{}}ResizedCrop (0.8,1.0), \\ HorizontalFlip, \\ VerticalFlip\end{tabular} & \begin{tabular}[c]{@{}c@{}}ResizedCrop (0.2,1.0),\\ HorizontalFlip\end{tabular} \\
Batch size    & 256                                                                                               & 256                                                                             \\
Learning rate & 1e-3                                                                                              & 0.1                                                                             \\
LR schedule   & cos                                                                                               & step                                                                            \\
Optimizer     & AdamW                                                                                             & SGD                                                                             \\
Weight decay  & 0.01                                                                                              & 0                                                                               \\
Warm-up       & 0                                                                                                 & 0                                                                               \\
Epoch         & 50                                                                                                & 100                                                                             \\ \hline
\end{tabular}

\end{table*}

\begin{table*}[t]
\caption{Transfer learning hyperparameters on the four GEO-Bench datasets.}
\label{tab:hyper-geobench}
\centering
\begin{tabular}{lcccc}
\hline
              & m-so2sat                                                                        & m-brick-kiln                                                                    & m-cashew-plantation                                                                                       & m-SA-crop-type                                                                                            \\ \hline
Backbone      & ViT-B/14, ViT-S/14   & ViT-B/14, ViT-S/14 & ViT-B/14, ViT-S/14 & ViT-B/14, ViT-S/14 \\
Input size    & 224x224                                                                         & 224x224                                                                         & 224x224                                                                                                   & 224x224                                                                                                   \\
Augmentation  & \begin{tabular}[c]{@{}c@{}}ResizedCrop (0.8,1.0),\\ HorizontalFlip\end{tabular} & \begin{tabular}[c]{@{}c@{}}ResizedCrop (0.8,1.0),\\ HorizontalFlip\end{tabular} & \begin{tabular}[c]{@{}c@{}}ResizedCrop (0.6,1.0),\\ HorizontalFlip,\\ VerticalFlip,\\ Rotate\end{tabular} & \begin{tabular}[c]{@{}c@{}}ResizedCrop (0.6,1.0),\\ HorizontalFlip,\\ VerticalFlip,\\ Rotate\end{tabular} \\
Batch size    & 256                                                                             & 256                                                                             & 64                                                                                                        & 64                                                                                                        \\
Learning rate & 1.0                                                                        & 10                                                                        & 0.01                                                                                                  & 0.01                                                                                                  \\
LR schedule   & cos                                                                             & cos                                                                             & cos                                                                                                       & cos                                                                                                       \\
Optimizer     & LARS                                                                            & LARS                                                                            & AdamW                                                                                                     & AdamW                                                                                                     \\
Weight decay  & 0                                                                               & 0                                                                               & 0.01                                                                                                      & 0.01                                                                                                      \\
Warm-up       & 0                                                                               & 0                                                                               & 3                                                                                                         & 3                                                                                                         \\
Epoch         & 50                                                                              & 50                                                                              & 50                                                                                                        & 50                                                                                                        \\ \hline
\end{tabular}

\end{table*}

\newpage

 

\begin{IEEEbiography}
[{\includegraphics[width=1in,height=1.25in,clip,keepaspectratio]{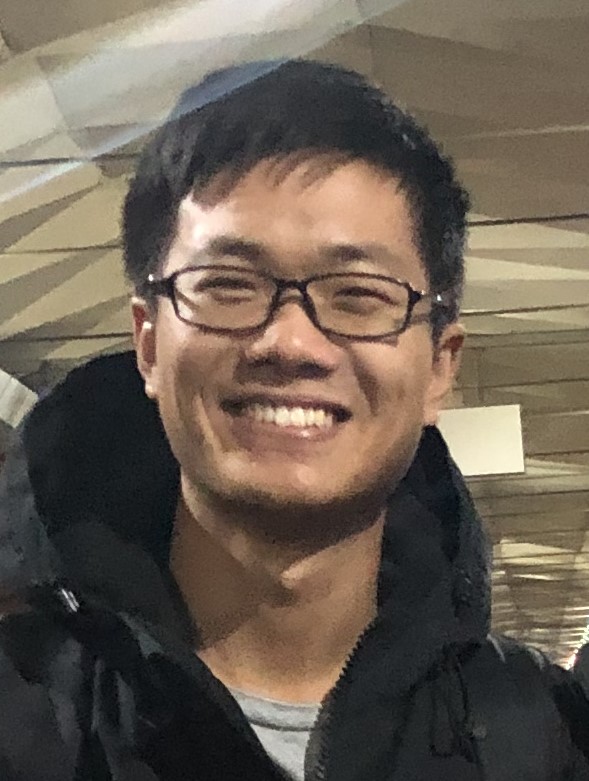}}] {Yi Wang} (S'21) received his B.E. degree in remote sensing science and technology from Wuhan University, Wuhan, China, in 2018 and his M.Sc. degree in geomatics engineering from University of Stuttgart, Stuttgart, Germany, in 2021. He is pursuing his Ph.D. degree at the Technical University of Munich (TUM), Munich, Germany. From 2021 to 2024, he was a research associate at the Remote Sensing Technology Institute, German Aerospace Center (DLR). In 2020, he spent three months at the perception system group, Sony Corporation, Stuttgart, Germany. His research interests include self-supervised learning, weakly-supervised learning, and multimodal representations. 
\end{IEEEbiography}

\begin{IEEEbiography}
[{\includegraphics[width=1in,keepaspectratio]{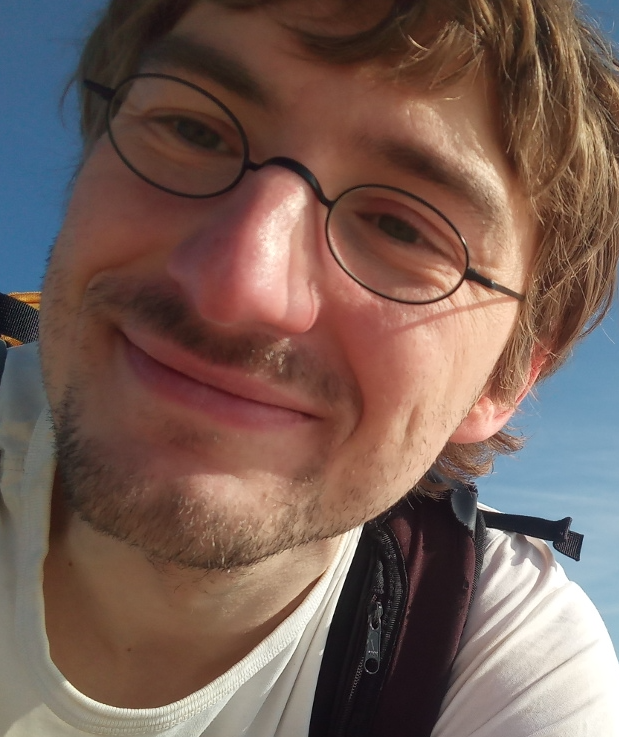}}] {Conrad M Albrecht} (M'17) is a researcher at the Earth Observation Center of the German Aerospace Center. Since April 2021, he is PI of the HelmholtzAI Young Investigator Group (YIG) "Large-Scale Data Mining in Earth Observation" (DM4EO). In July 2023, he was appointed a visiting associate professor with the Institute of Nasca at Yamagata University, Japan contributing to research in machine learning for the UNESCO World Heritage of the Nasca culture in Peru.

For over 6 years Conrad was a research scientist in the Physical Sciences department at the IBM T.J. Watson Research Center in Yorktown, NY, USA. While at the Institute for Theoretical Physics, he graduated in physics (International Max-Planck Research School for Quantum Dynamics in Physics, Chemistry and Biology) with an extra certification in computer science (Cluster- \& Detector Management team at CERN, Switzerland). He received a corresponding Ph.D. degree from Heidelberg University, Germany in 2014 working on distributed computing to study physics at low temperatures.

Conrad's research agenda interconnect physical models and numerical analysis, employing Big Data technologies and machine learning through open-science research, \url{https://conrad-m-albrecht.github.io}. Conrad co-organized workshops at the IEEE BigData conference, IGARSS conferences, and the AAAS annual meeting. Conrad is home in Europe and the United States. Some of his transatlantic initiatives include: In 2023, he was invited by the Alexander-von-Humboldt Foundation to present at the German-American Frontiers of Engineering Symposium. Beginning in 2024, initiated by Helmholtz Imaging and Data Science, he is in close touch with the German Department at Princeton University for cross-atlantic undergraduate internships. For fall 2024, Conrad was invited Adjunct Professor of Earth and Environmental Engineering by Columbia University, NY, USA.
\end{IEEEbiography}

\begin{IEEEbiography}
[{\includegraphics[width=1in,height=1.25in,clip,keepaspectratio]{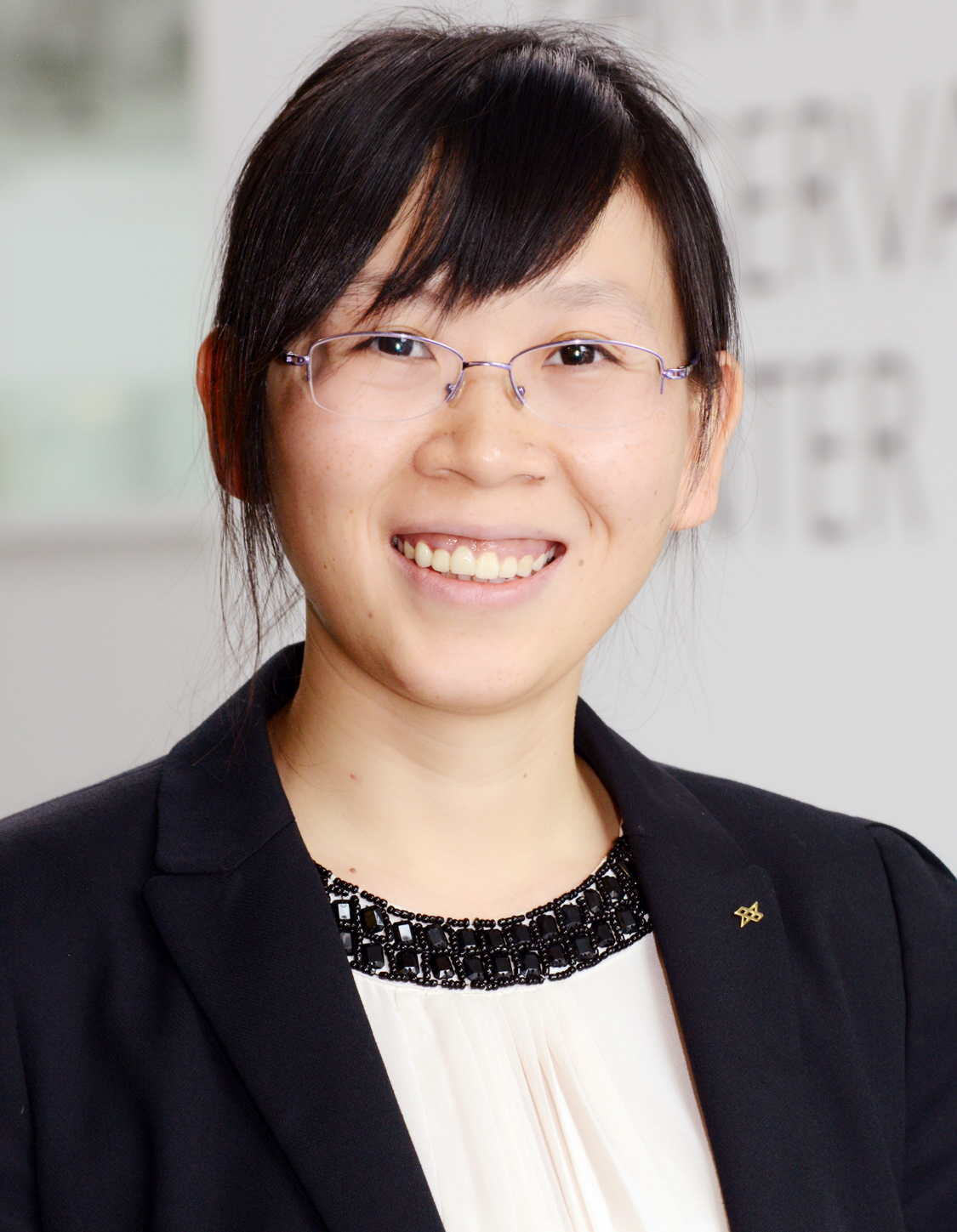}}]{Xiao Xiang Zhu}(S'10--M'12--SM'14--F'21) received the Master (M.Sc.) degree, her doctor of engineering (Dr.-Ing.) degree and her “Habilitation” in the field of signal processing from Technical University of Munich (TUM), Munich, Germany, in 2008, 2011 and 2013, respectively.
\par
She is the Chair Professor for Data Science in Earth Observation at Technical University of Munich (TUM) and was the founding Head of the Department ``EO Data Science'' at the Remote Sensing Technology Institute, German Aerospace Center (DLR). Since May 2020, she is the PI and director of the international future AI lab "AI4EO -- Artificial Intelligence for Earth Observation: Reasoning, Uncertainties, Ethics and Beyond", Munich, Germany. Since October 2020, she also serves as a Director of the Munich Data Science Institute (MDSI), TUM. From 2019 to 2022, Zhu has been a co-coordinator of the Munich Data Science Research School (www.mu-ds.de) and the head of the Helmholtz Artificial Intelligence -- Research Field ``Aeronautics, Space and Transport".  Prof. Zhu was a guest scientist or visiting professor at the Italian National Research Council (CNR-IREA), Naples, Italy, Fudan University, Shanghai, China, the University  of Tokyo, Tokyo, Japan and University of California, Los Angeles, United States in 2009, 2014, 2015 and 2016, respectively. She is currently a visiting AI professor at ESA's Phi-lab, Frascati, Italy. Her main research interests are remote sensing and Earth observation, signal processing, machine learning and data science, with their applications in tackling societal grand challenges, e.g. Global Urbanization, UN’s SDGs and Climate Change.

Dr. Zhu has been a member of young academy (Junge Akademie/Junges Kolleg) at the Berlin-Brandenburg Academy of Sciences and Humanities and the German National  Academy of Sciences Leopoldina and the Bavarian Academy of Sciences and Humanities. She is a Fellow of the Academia Europaea (the Academy of Europe). She serves in the scientific advisory board in several research organizations, among others the German Research Center for Geosciences (GFZ, 2020-2023) and Potsdam Institute for Climate Impact Research (PIK). She is an associate Editor of IEEE Transactions on Geoscience and Remote Sensing, Pattern Recognition and served as the area editor responsible for special issues of IEEE Signal Processing Magazine (2021-2023). She is a Fellow of IEEE, AAIA, and ELLIS.
\end{IEEEbiography}



\vfill

\end{document}